\documentclass[11pt]{article}
\usepackage[final]{acl}

\usepackage{times}
\usepackage{latexsym}
\usepackage[T1]{fontenc}
\usepackage[utf8]{inputenc}
\usepackage{microtype}
\usepackage{inconsolata}
\usepackage{graphicx}
\usepackage{booktabs}
\usepackage{array}
\usepackage{amsmath,amssymb}
\usepackage{float}
\usepackage{parskip}

\newcommand{\code}[1]{\texttt{#1}}
\newcommand{\artifact}[1]{\nolinkurl{#1}}

\usepackage{xparse}
\usepackage{tabularx}
\usepackage{array}

\ExplSyntaxOn

\NewDocumentCommand{\PromptBox}{m}
  {
    \par
    \fbox{%
      \begin{minipage}{%
        \dimexpr\linewidth-2\fboxsep-2\fboxrule\relax
      }
        \begin{tabularx}{\linewidth}
          {@{}>{\bfseries}l@{\;}>{\itshape\arraybackslash}X@{}}
          \clist_map_function:nN {#1} \promptbox_row:n
        \end{tabularx}
      \end{minipage}%
    }
    \par
  }

\cs_new_protected:Nn \promptbox_row:n
  {
    \promptbox_row_aux:nn #1
  }

\cs_new_protected:Nn \promptbox_row_aux:nn
  {
    #1 & #2 \\
  }

\ExplSyntaxOff

\title{Dynamic Lagging using Stable-Prefix Training for Simultaneous Translation}
\author{Hieu Hoang \\
  Microsoft\\
  \And
  Amittai Axelrod
  \And
  Matt Post \\
  Microsoft}
\begin{document}
\maketitle

%%%%%%%%%%%%%%%%%%%%%%%%%%%%%%%%%%%%%%%%%%%%%%%%%%
\begin{abstract}
    In streaming simultaneous speech translation, the speech translation system is trained to learn a read-write policy that alternates between consuming source words and generating target ones.
    In a cascaded setting, the output from the speech recognizer is passed to a separate machine translation component, making it more difficult to learn such a policy.
    Approximations such as fixed wait-$k$ strategies or target-suffix deletion can be employed, but these approaches do not provide the model with a streaming system's flexibility to make contextual read-write decisions.
    This paper presents a training strategy for a cascaded machine translation system that enables it to dynamically decide how much of the growing source prefix to translate.
    We achieve this by fine-tuning a large language model (Qwen3-8B) on \emph{stable prefixes} of the training data, which are produced by pairing every source sentence prefix in the training data with the longest translation of that prefix that is shared with the full source sentence translation.
    We fine-tune variants of the model on different subsets of the prefixes and compare against wait-$k$ and target-suffix deletion.
    We also investigate the effect of fine-tuning the target-token generation confidence.
    Our experiments show that stable prefixes improve the quality-latency tradeoff when translating from English into German, Japanese, and Chinese across a range of test sets.

\end{abstract}

%%%%%%%%%%%%%%%%%%%%%%%%%%%%%%%%%%%%%%%%%%%%%%%%%%
\section{Introduction}

Simultaneous speech translation begins translating before the complete speech input has been observed.
In this setting, there is a natural tradeoff between translation quality (which is highest when the entire input is available) and latency (which is lowest when the system begins translating right away).
A number of approaches have been developed to address this tradeoff, including fixed schedules such as wait-$k$ \cite{ma-etal-2019-stacl}, adaptive policies that make the read-write decision dynamically \cite{gu-etal-2017-learning,zheng-etal-2020-simultaneous}, re-translation \citep{arivazhagan-etal-2019-retranslation}, or local agreement \citep{polak-etal-2022-cuni}.
% One set of approaches has been to develop new training and inference algorithms that couple the speech recognition and translation into a single system that switches between consuming source words from the input and generating target words for the translation.

A cascaded pipeline, where the text output of an automatic speech recognition (ASR) system is passed to a machine translation (MT) system, continues to be an effective approach \citep{bentivogli-etal-2021-cascade}.
For the machine translation component, a cascade can employ a specialized streaming architecture or make use of off-the-shelf, standard translation systems or large language models, which can also be tuned or adapted in different ways.
The machine translation system must determine how much of the input it will translate, but for standard architectures, the options are often limited to wait-$k$ or target-suffix deletion (masking), which are coarse policies that are not responsive to the source context.
% An alternative is to use a cascade, in which an automatic speech recognition (ASR) system is paired with a machine translation (MT) system in a pipeline.
% In this setting, the translation component does not have control over how much input it receives, but can only decide how much of that input to translate.
% Higher quality can be obtained by retranslating from scratch, but this produces \emph{flicker} in the translation output as previous translations are overwritten.
% Instead, the MT system can commit to the generated output, using prefix-constrained decoding.
% This eliminates flicker, but may affect quality, especially if the commitment is made too early, when only a short input prefix is available.
% The MT system can reassert some control using source- or target-suffix deletion, but this is a coarse tool that is not responsive to context.

In this paper, we train an MT system to dynamically decide how much output it is willing to commit to.
We do this by introducing the idea of a \emph{stable translation prefix}, a translation of a source prefix that has been truncated for consistency with the full-sentence translation.
This is computed by first using a trained MT system to translate the source side of every sentence in the training data, as well as every prefix of those sentences, and then pairing each source prefix with the longest common prefix between its translation and the full-sentence translation.
These pairs then constitute training data that can be used for fine-tuning, allowing the model to learn how much to ``trust'' any particular source prefix.
% These then allow us to fine-tune a translation model that learns how much of a target translation it should commit to, automatically trading off quality and latency.

Our experiments demonstrate the effectiveness of this approach.
In experiments fine-tuning Qwen3-8B, we observe substantial improvements in the latency-quality Pareto frontier for models trained on these prefixes and evaluated with wait-$k$, target-suffix deletion, and a tuned confidence threshold.
These findings hold across three language pairs (en-de, en-ja, and en-zh) on FLEURS \cite{conneau-etal-2022-fleurs}, WMT24++ \cite{deutsch-etal-2025-wmt24}, and CoVoST 2 \cite{wang-etal-2020-covost2} test sets.

\section{Background}

Simultaneous machine translation begins translating before the entire input is received.
In the streaming setting \citep{cho-esipova-2016-can}, the output, once generated, cannot be revised, and the system's core task is to determine whether to read the next source token or generate a target token.\footnote{We follow \citet{arivazhagan-etal-2020-translation} in distinguishing streaming from re-translation approaches, in which the output can be changed.}
% In simultaneous translation, theWith end-to-end models \citep{cho-esipova-2016-can, huang-etal-2020-simultaneous}, a single system is trained, and its core task is deciding whether to read the next source token or write a target token.
% Here, the core task is deciding whether to read the next source token or write a target token.
Early work learned this policy directly, either with reinforcement learning \cite{gu-etal-2017-learning} or with supervised agents trained from oracle decisions \cite{zheng-etal-2020-simultaneous}. 
Fixed policies offer a simpler alternative. 
In particular, wait-$k$ reads $k$ source tokens and then alternates between reading and writing \cite{ma-etal-2019-stacl}, providing an explicit latency-quality control but without considering the context.
These transduction approaches commit target tokens incrementally.
This formulation is naturally suited to simultaneous translation but requires new training and inference paradigms. 

Cascaded approaches, which feed the output of an ASR system into an MT one, remain popular \citep{bentivogli-etal-2021-cascade}.
% In this setting, the output of an automatic speech recognition (ASR) system is passed to a text-based MT system. 
This setting permits the use of specialized encoder-decoder models \cite{huang-etal-2020-simultaneous}, but it is also possible to use standard translation models which either translate input chunks independently, or which repeatedly re-translate the source, constrained each time to the previously generated target prefix \citep{polak-etal-2022-cuni}.
% Standard translation models, trained in the usual fashion on sentence pairs, can also be employed for this task, for example continually retranslating the source from the beginning \citep{niehues-etal-2016-dynamic}.
% A downside is that repeated retranslation causes instability in the output, which often changes as the string is retranslated.
% This can be mitigated by using forced translation (e.g., \citet{polak-etal-2022-cuni}), which constrains each retranslation to be prefixed by the previous translation, or with self-training on partial prefixes \cite{sen-etal-2023-self}.

A key distinction for the MT component is whether the architecture is the ``standard'' model (trained to consume the entire input and produce an output) or has been adapted for streaming (and has been trained to use or dynamically decide a customized read-write policy).
Standard models are limited in their ability to control the amount of output text.
One approach is to impose a wait-$k$ policy on top of the model.
Another is to use ``target suffix deletion (mask-$k$)'' \cite{arivazhagan-etal-2019-retranslation}, which removes words from the end of the partial translation. 
This allows the caller to wait until the next, longer partial source without committing to the complete partial translation, at the cost of some latency, and it reflects the fact that translation uncertainty falls as more of the source is observed.
However, both these approaches are coarse controls that do not take into account the actual uncertainty or stability of the translation at each point in the source.
Our approach is to train the model on dynamically-determined stable prefixes and to use a tuned confidence threshold.

% Prior work has looked at ways of modifying the training data for partial speech translation.
% For example, \citet{ma-etal-2019-stacl} propose prefix-to-prefix training, where the model is trained on source prefixes and the corresponding truncated reference translations.
% \citet{sen-etal-2023-self} explore self-training on partial prefixes, which allows the model to learn from its own predictions on incomplete source segments.
% As far as we know, our work is distinct in producing \emph{stable} prefixes, which are the longest common prefixes between the translation of a source prefix and the translation of the full source sentence.

With the high performance of LLM-based translation systems, it is natural to incorporate them into speech translation pipelines. 
\citet{yang-etal-2025-large-language} surveys the integration of the speech modality into (largely text-based) LLMs. 
A number of papers construct training data that is amenable to multi-turn dialog \cite{ouyang-etal-2025-infinisst, wang-etal-2025-conversational}. \citet{agostinelli-etal-2024-simul} provide a fine-tuning and evaluation framework for LLMs, and \citet{fu-etal-2025-llms} train on interleaved source-target pairs under latency constraints.
This paper uses Qwen3-8B \citep{qwen2024} as a base translation model for evaluation and fine-tuning.

Prefixes have been used in various ways for training and inference.
In prefix training \citep{niehues-etal-2016-dynamic}, each (source, target) training pair is randomly truncated.
\citet{sen-etal-2023-self} explore self-training on tagged prefix training data, demonstrating increased monotonicity and reduced flicker in a re-translation setting.
% As far as we know, our work is distinct in producing \emph{stable} prefixes, which are the longest common prefixes between the translation of a source prefix and the translation of the full source sentence.
At inference time, \citet{polak-etal-2022-cuni} compared hypotheses in the beam, and found the best performance from ``local agreement'', which computes a target prefix from a sequence of translations of growing source prefixes. 
The local agreement criterion itself was introduced by \citet{liu-etal-2020-low}, who commit the longest prefix shared by consecutive hypotheses. 
The same principle of committing only the stable portion of a hypothesis underlies re-translation systems that stabilize the output prefix as more source arrives~\citep{niehues-etal-2016-dynamic,niehues-etal-2018-low,arivazhagan-etal-2019-retranslation}, and agreement between successive prefix translations was already exploited as a commit criterion by \citet{cho-esipova-2016-can}. 
Our approach extends work in prefix training with the idea of a ``stable'' prefix: a distilled translation prefix anchored to the full-sentence translation.
% Our work differs from these inference-time uses by using the stable prefix to create finetuned training data, computing it relative to the unconstrained translation of the complete source. 
% Unlike prefix-to-prefix training~\citep{ma-etal-2019-stacl}, which truncates the reference at a fixed wait-$k$, our supervision is derived from the model's own agreement with its offline, sentence-level, hypothesis.

%%%%%%%%%%%%%%%%%%%%%%%%%%%%%%%%%%%%%%%%%%%%%%%%%%
\section{Approach}

\subsection{Prefix Translation Synthesis}
\label{section:stable}

\begin{table*}[t]
    \centering
    \begin{tabular}{llp{1.8in}p{1.6in}p{1.6in}}
    \toprule
    \textbf{\#} & \textbf{type} & \textbf{source prefix} & \textbf{translation} & \textbf{stable prefix} \\
    \midrule\midrule
    & final & Does he playfully tease you? & Neckt er dich spielerisch? & Neckt er dich spielerisch? \\
    \midrule
    1 & partial & Does & Macht \\
    2 & partial & Does he & Tut er das \\
    3 & partial & Does he playfully & Tut er das spielerisch \\
    4 & partial & Does he playfully tease & Neckt er & Neckt er \\
    \bottomrule
    \end{tabular}
\caption{Stable-prefix extraction for a German example. Not until step 4 is a partial translation a prefix of the full-sentence translation.}
    \label{table:stable-prefix}
\end{table*}

Our goal is to produce a system which, when presented with a source prefix, will only translate as much of it as it can do so reliably. 
Past work in prefix training relies on randomly sampling the prefix, which does not properly account for prefix ambiguity.
We address that by constructing training data based on the notion of a \textbf{stable prefix}.

Given a monolingual source sentence $s$ and a translation system $t$, we produce a translation $f=t(s)$. 
We also compute $p_i=t(s_{1..i})$, where each $s_{1..i}$ contains the first $i$ words of $s$. 

Let $\ell_i=|\operatorname{lcp}(p_i,f)|$ be the length of their longest common word-level (EN$\rightarrow$DE) or character-level prefix (EN$\rightarrow$JA and EN$\rightarrow$ZH).

For source prefix $s_{1..i}$, we define the \emph{stable prefix} as $p'_i=f_{1..m_i}$, where $m_i=\max_{j\le i}\ell_j$. 

Thus the target prefix grows monotonically: once content has agreed with the full-source translation and been committed, a noisier later partial translation cannot retract it. 
This increases the prefix translation length and stability of the training data. 
The pseudocode for this procedure can be found in Appendix~\ref{sec:prefix-algorithms}.

The stable-prefix translation corpus is merged with the full translations when finetuning the model at a fixed ratio.
Table~\ref{table:stable-prefix} contains an example. 
As we translate successively longer source prefixes, the translations in the ``translation'' column exhibit ``jitter.'' 
Any suffix of the partial translation beyond the stable prefix is unstable and is removed. 

\subsection{LLM Prompting}
\label{sec:LLM Prompting}

The LLM is prompted to generate translations of a source prefix. It is told that the input is a partial source sentence; its input and output are not constrained or truncated.
New source arriving incrementally is translated by appending it to the previously observed source and re-prompting the model to translate in the same manner. The previously committed target is supplied as a forced prefix. 
An incremental input that completes the source sentence is also force-decoded with the previously committed translation, but the model is explicitly told it is a complete sentence.
The architecture is flicker-free by construction, as the model has the previous translation as context but cannot change it.

Consider the source sentence example \emph{Scientists discovered a new species in the rainforest}, where the model assistant has already committed a translation for the prefix \emph{Scientists discovered a new}:
\PromptBox{ {User$_1$}{Scientists discovered a new}, {Asst.$_1$}{Wissenschaftler entdeckten eine neue},
}
When the next chunk \emph{species} arrives---the sentence is still incomplete---the model is re-prompted with the source observed so far and the committed target as a forced prefix, and emits only the new continuation:
\PromptBox{ {User$_2$}{Scientists discovered a new species}, {Forced$_2$}{Wissenschaftler entdeckten eine neue}, {Asst.$_2$}{Art}
}
Finally the chunk \emph{in the rainforest} completes the sentence; the model is re-prompted with the full source---now flagged as complete---and the committed target as a forced prefix, and emits the remaining continuation:
\PromptBox{ {User$_3$}{Scientists discovered a new species in the rainforest}, {Forced$_3$}{Wissenschaftler entdeckten eine neue Art}, {Asst.$_3$}{im Regenwald}
}
%

%%%%%%%%%%%%%%%%%%%%%%%%%%%%%%%%%%%%%%%%%%%%%%%%%%
%===================================================
\section{Experimental setup}
\label{sec:Experimental setup}

\subsection{Models}

We use Qwen3-8B \cite{qwen2024} throughout this paper.

\subsection{Data}

A common practice in MT and LLM training is to use a more capable model to generate distilled data for a less capable model. We take a self-distillation approach, using a single model to generate synthetic parallel data and then fine-tuning on this data to make the model prefix-aware for streaming translation. 

Self-distillation also provides a soft upper bound on performance, enabling us to compare the fine-tuned model with the original model that generated the data. 
Distilled data is also more regular than raw parallel data, so more prefix examples can be extracted to enrich the training data for the prefix-aware model.

We experiment in three language directions: EN$\rightarrow$DE, EN$\rightarrow$JA, and EN$\rightarrow$ZH. The source sentences are a randomly sampled subset of the WMT24 training data \citep[EN$\rightarrow$DE]{kocmi-etal-2024-findings} and WMT25 training data \citep[EN$\rightarrow$JA and EN$\rightarrow$ZH]{kocmi-etal-2025-findings}. 

20 million parallel full sentences are synthesized as well as their corresponding prefix pairs for each language direction using the base LLM. The stable prefixes are then generated from this data as described in Section~\ref{section:stable}.

Validation is performed on held-out data (newstest2019~\citep{barrault-etal-2019-findings} for EN$\rightarrow$DE and wmttest2023~\citep{kocmi-etal-2023-findings} for EN$\rightarrow$JA and EN$\rightarrow$ZH) for full sentence pairs. 
We also validate on stable prefix pairs held out from the training data along with their full sentence pairs.

The models are evaluated on WMT24++ \cite{deutsch-etal-2025-wmt24}, FLEURS
\cite{conneau-etal-2022-fleurs}, and CoVoST~2 \cite{wang-etal-2020-covost2} using COMET-22-DA \cite{rei-etal-2022-comet} and MetricX-24-Hybrid-XL \cite{juraska-etal-2024-metricx}.

\subsection{Training and Validation}
We finetune two models, \code{prefix} and \code{continuation}, both using LoRA adapters \cite{hu-etal-2022-lora} with default settings and the Adam optimizer \citep{kingma-ba-2015-adam}, with a learning rate of $5\times10^{-6}$ and an effective batch size of 128.

Training follows the force-decoding paradigm described in Section~\ref{sec:LLM Prompting}; non-initial prompts are given the source translation from the beginning of the sentence and force-decode with the previous target prefix. Loss is applied only to the continuation and end-of-sequence tokens.

The different training strategies for the \code{prefix} and \code{continuation} models are described in the next two sections.

\subsubsection{\code{prefix} model}
\label{sec:bootstrap-monotonic-fd}

\paragraph{Training}
draws an equal mixture of prefix and full-sentence examples and trains them as cross-entropy tasks. Samples from the prefix corpus are prompted with the initial translation prompt; full-sentence pairs use the final translation prompt.  Thus the \code{prefix} model learns the initial-prefix and initially-complete boundary cases, but it receives no examples containing an already committed target prefix and no intermediate continuation examples. 

\paragraph{Validation}
occurs every 500 updates and measures COMET on the full-sentence development set and on prefix pairs.  Prefix COMET is first averaged within each observed-source-length bin and then equally across bins; the selection score is the equally weighted mean of this prefix score and full-sentence COMET. Training stops when the validation score does not improve for 5 consecutive epochs; the highest-selection-score checkpoint is then used for evaluation.

\subsubsection{\code{continuation} model}
\label{sec:continuation}

\paragraph{Training}
is initialised from the \code{prefix} model. Whereas \code{prefix} is trained only on the initial-prefix and full-sentence boundary cases, the \code{continuation} model adds two similar tasks; \emph{continue} translates a chunk of words where its prefix has already been translated, and \emph{final-continue} translates the final chunk of the sentence given an already-committed target prefix. These four tasks are sampled during training with the following probabilities: 
\begin{itemize}
\item \emph{initial} 60\%
\item \emph{continue} 20\%
\item \emph{final-continue} 10\%
\item \emph{initially-complete} 10\%
\end{itemize}

\paragraph{Validation and checkpoint selection.}
We hold out a fixed slice of 64 training sentences. On each held-out sentence we imitate inference: the source is revealed a few words at a time, and at every step the model continues from \emph{its own} previously committed text (using the same force-decoding prompt as training, never the gold prefix); these are then concatenated into one complete translation of the sentence. COMET is computed for this incrementally built translation, and for translating the whole sentence at once. The selection score is the equally weighted average of these two COMET scores. Training stops when the validation score does not improve for 5 consecutive epochs.

\subsection{Evaluation}

We apply two evaluation methods: binary-segmentation translation and incremental translation.

In the first case, we segment the source sentences at every word boundary into two segments and translate the prefix and the suffix. The prefix translation is used as context, via force decoding, to translate the suffix. The COMET score \cite{rei-etal-2022-comet} is computed over the concatenated prefix and suffix translations. This reveals how translation quality changes depending on where the source is segmented.

The incremental evaluation is more realistic, as it simulates source text arriving incrementally in chunks of different lengths, reflecting the conditions of a live simultaneous translation scenario. At each step, the model has to decide whether---and how much---to commit to the translation output. 
We use Average Lagging \cite{ma-etal-2019-stacl} to measure how long a system waits before producing translation output: for a value $k$, the system produces the translation only up to input token $n-k$.

%%%%%%%%%%%%%%%%%%%%%%%%%%%%%%%%%%%%%%%%%%%%%%%%%%
\section{Results}

\subsection{Finetuning LLM}

\begin{table}[t]
\centering
\footnotesize
\caption{Average Lagging and COMET ($\times100$) on FLEURS. \emph{Full} is complete-sentence COMET; \emph{Worst} is the minimum corpus-average COMET over source boundaries.}
\label{tab:al-comet-sft}
\setlength{\tabcolsep}{4pt}
\begin{tabular}{@{}rlrrr@{}}
\toprule
ID & model & AL & Full & Worst \\
\midrule
% BEGIN CLEANBINARY_TABLE
\multicolumn{5}{@{}l}{\textbf{EN$\rightarrow$DE}} \\
1 & \code{base} & \textbf{0.82} & 85.9 & 80.9 \\
2 & \code{prefix} & 2.21 & \textbf{87.1} & \textbf{85.1} \\
3 & \code{continuation} & 2.21 & 86.1 & 81.2 \\
\midrule
\multicolumn{5}{@{}l}{\textbf{EN$\rightarrow$JA}} \\
4 & \code{base} & \textbf{0.11} & 89.2 & 81.3 \\
5 & \code{prefix} & 3.06 & \textbf{90.4} & \textbf{90.0} \\
6 & \code{continuation} & 1.06 & 89.2 & 75.7 \\
\midrule
\multicolumn{5}{@{}l}{\textbf{EN$\rightarrow$ZH}} \\
7 & \code{base} & 0.60 & 88.1 & 85.8 \\
8 & \code{prefix} & 3.11 & \textbf{88.8} & \textbf{88.4} \\
9 & \code{continuation} & \textbf{0.55} & 88.3 & 85.5 \\
% END CLEANBINARY_TABLE
\bottomrule
\end{tabular}
\end{table}

\begin{figure*}[!t]
\centering
\begin{minipage}{0.28\textwidth}
  \centering
\textbf{EN$\rightarrow$DE}\\[2pt]
  \includegraphics[width=\linewidth]{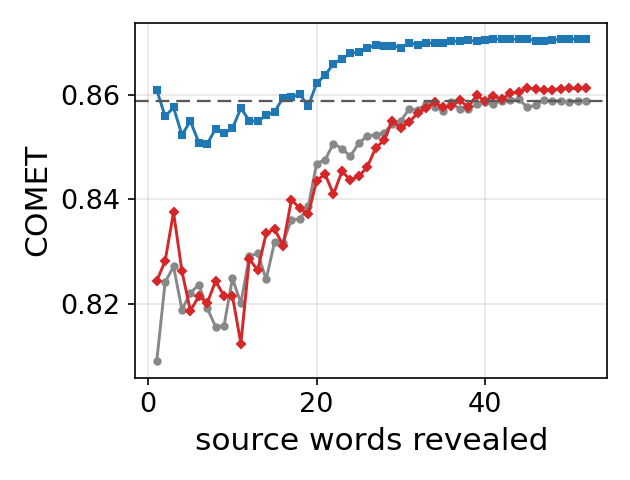}
\end{minipage}\hfill
\begin{minipage}{0.28\textwidth}
  \centering
\textbf{EN$\rightarrow$JA}\\[2pt]
  \includegraphics[width=\linewidth]{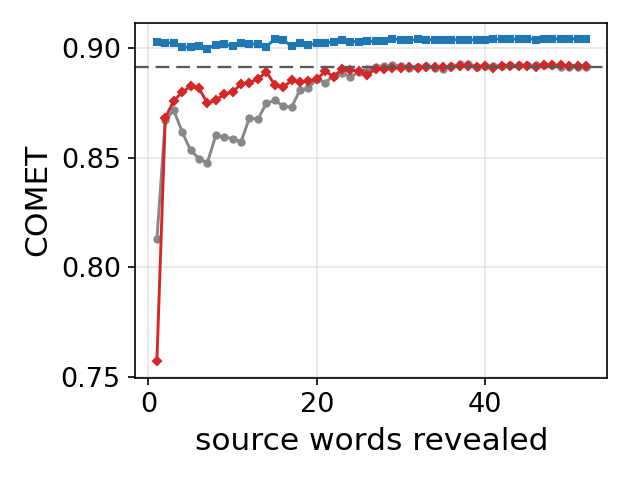}
\end{minipage}\hfill
\begin{minipage}{0.28\textwidth}
  \centering
\textbf{EN$\rightarrow$ZH}\\[2pt]
  \includegraphics[width=\linewidth]{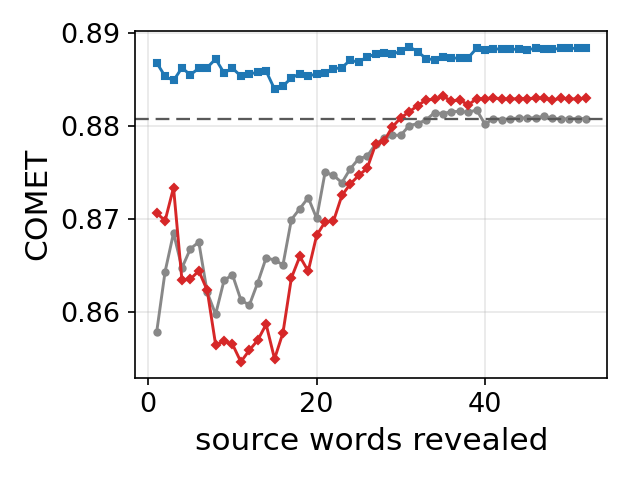}
\end{minipage}

\vspace{6pt}
\begin{minipage}{0.28\textwidth}
  \centering
  \includegraphics[width=\linewidth]{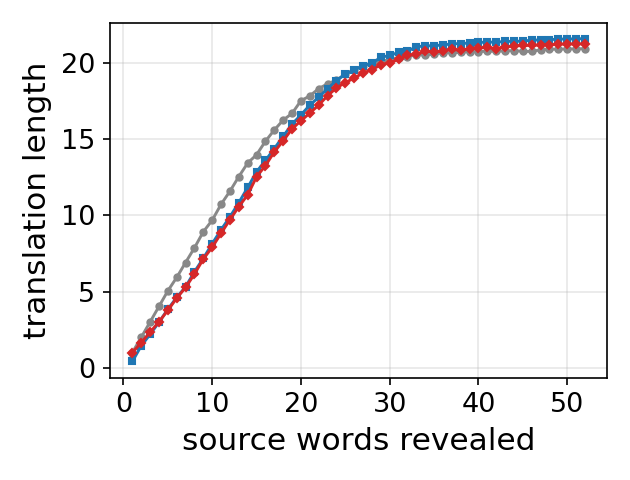}
\end{minipage}\hfill
\begin{minipage}{0.28\textwidth}
  \centering
  \includegraphics[width=\linewidth]{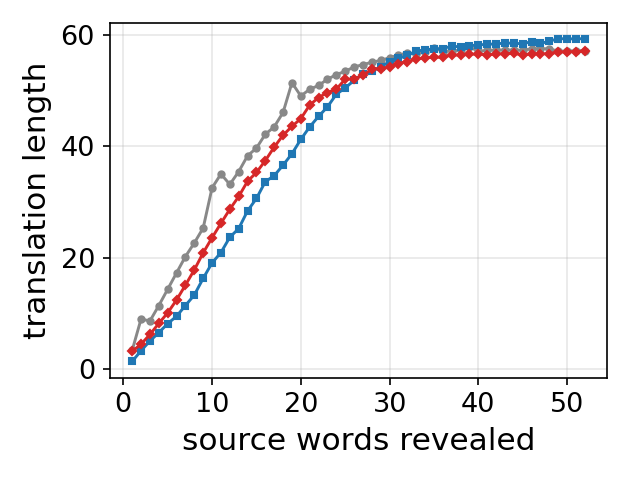}
\end{minipage}\hfill
\begin{minipage}{0.28\textwidth}
  \centering
  \includegraphics[width=\linewidth]{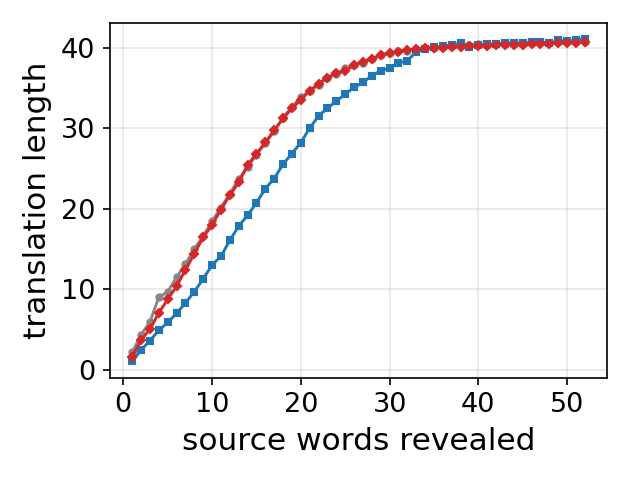}
\end{minipage}

\vspace{2pt}
\includegraphics[width=0.6\linewidth]{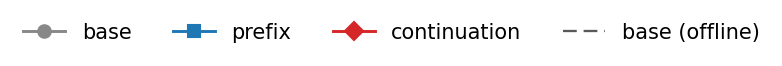}
\caption{Binary source-completion on FLEURS: COMET of the recomposed translation (top) and prefix-translation length (bottom) versus source words revealed, for the pre-trained \code{base} and the prefix adapter \code{prefix}.}
\label{fig:stateful-bootstrap-binary}
\end{figure*}

The results of the binary segmentation evaluation are summarized in Table~\ref{tab:al-comet-sft} and visualized in Figure~\ref{fig:stateful-bootstrap-binary} for the \code{base} (pre-trained LLM), \code{prefix}, and \code{continuation} models. We measured average lag and the COMET score on the test set when translating full sentences. 

We also calculated the COMET score at each segmentation position. If a particular sentence is shorter than the segmentation point, its score is taken as the COMET score of the full sentence. The \emph{Worst} score in Table~\ref{tab:al-comet-sft} corresponds to the lowest COMET score observed across all segmentation positions for each model.

We note a few interesting observations:
\begin{itemize}
\item The prefix models translate full sentences better according to COMET, even better than the base model from which the distilled data came.  This resembles the ``born-again'' phenomenon, in which a student trained on a teacher's predictions surpasses that teacher despite sharing its architecture, because self-distillation can regularize and reweight the learning signal~\citep{furlanello-etal-2018-born}; here, the large ratio of full-sentence examples likely strengthens that effect. Continuation matches the pre-trained model.

\item The \code{base} model is most aggressive at committing translations. The average lag of the \code{prefix} model is significantly higher (worse) than the other models, for all 3 language pairs. The bottom row of Figure~\ref{fig:stateful-bootstrap-binary} indicates the \code{prefix} model sees a lot more source tokens than the other models for a given quantity of output tokens.

\item Figure~\ref{fig:stateful-bootstrap-binary} shows that base-model translation quality dips at early word positions when source sentences are segmented, translated in two parts, and the translations recomposed. The \code{prefix} model is noticeably better than the \code{base} model in all three languages at the cost of higher latency. The \code{continuation} model is a little better than \code{base}.
\end{itemize}

\subsection{Pareto frontier}

The binary segmentation evaluation provides insight into how the models perform when translating source prefixes of varying lengths; however, it is not a realistic simulation of real-time translation scenarios, where the source unfolds incrementally in unpredictably sized chunks.

We simulate this scenario by incrementally revealing randomly sized source chunks to the models, with the chunk sizes drawn from a uniform distribution between 1 and 5. The average quality of the full, concatenated translation is measured against latency to gauge the trade-off between the two competing priorities for each model.

On their own, each of the models occupies one point in the quality-latency space. We now introduce three inference-time mechanisms to control the trade-off between translation quality and latency for any model, which lets us trace out a continuous Pareto frontier for each model.
\begin{enumerate}
\item \textbf{wait-\emph{k}}: the classical wait-$k$ policy with its own canonical read schedule---read the first $k$ source words, then alternate reading one source word and committing one target unit ($g(t)=\min(k{+}t{-}1,|x|)$). A larger $k$ waits for more source tokens before committing, thus raising latency.
\item \textbf{target suffix deletion}: at each non-final step the model generates its full continuation but 
deletes the last $d$ target units (words, or characters for EN$\rightarrow$JA/ZH) before committing. This 
holds back the least-reliable tail tokens from the output; larger $d$ is more conservative and adds latency.
\item \textbf{confidence}: a threshold $\tau$ on the stop-versus-continue margin between the probability of generating EOS (end of sentence) versus anything else. The action chosen is:
\begin{equation}
\begin{aligned}
\Delta &= P(\text{best non-EOS})-P(\text{EOS}),\\
a(\Delta;\tau) &=
\begin{cases}
\textsc{commit}, & \Delta\ge\tau,\\
\textsc{stop}, & \Delta<\tau.
\end{cases}
\end{aligned}
\end{equation}
$\tau{\to}{-}1$ commits aggressively, for low latency, while $\tau{\to}1$ waits for the whole source, recovering the offline translation. Using the model's own output confidence to drive the read/write decision follows a line of adaptive simultaneous-translation policies~\citep{cho-esipova-2016-can,gu-etal-2017-learning,zheng-etal-2019-simpler,zheng-etal-2020-simultaneous,ma-etal-2020-monotonic}.
\end{enumerate}

\begin{figure*}[!t]\centering
\begin{minipage}{0.32\linewidth}
  \centering
\textbf{\footnotesize EN$\rightarrow$DE}
\end{minipage}\hfill
\begin{minipage}{0.32\linewidth}
  \centering
\textbf{\footnotesize EN$\rightarrow$JA}
\end{minipage}\hfill
\begin{minipage}{0.32\linewidth}
  \centering
\textbf{\footnotesize EN$\rightarrow$ZH}
\end{minipage}

\vspace{2pt}
\textbf{\footnotesize \code{base}}\\[1pt]
\begin{minipage}{0.32\linewidth}
  \centering
  \includegraphics[width=\linewidth]{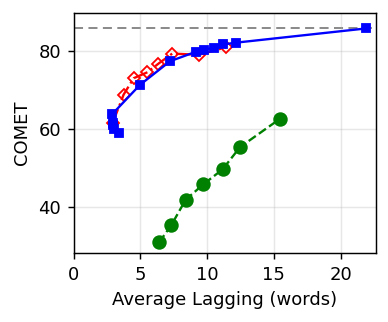}
\end{minipage}\hfill
\begin{minipage}{0.32\linewidth}
  \centering
  \includegraphics[width=\linewidth]{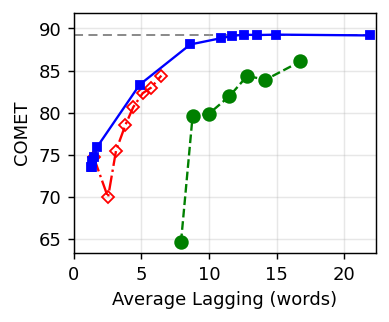}
\end{minipage}\hfill
\begin{minipage}{0.32\linewidth}
  \centering
  \includegraphics[width=\linewidth]{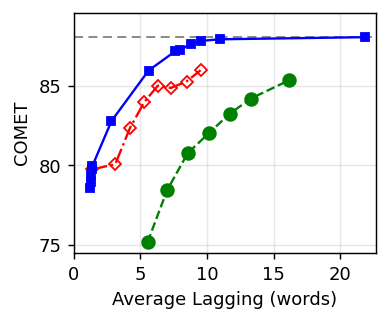}
\end{minipage}

\vspace{3pt}
\textbf{\footnotesize \code{prefix}}\\[1pt]
\begin{minipage}{0.32\linewidth}
  \centering
  \includegraphics[width=\linewidth]{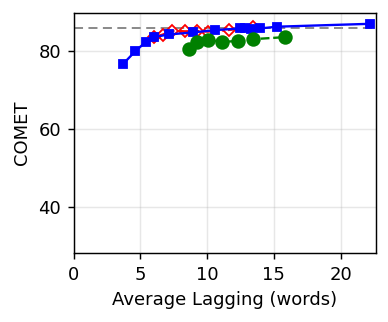}
\end{minipage}\hfill
\begin{minipage}{0.32\linewidth}
  \centering
  \includegraphics[width=\linewidth]{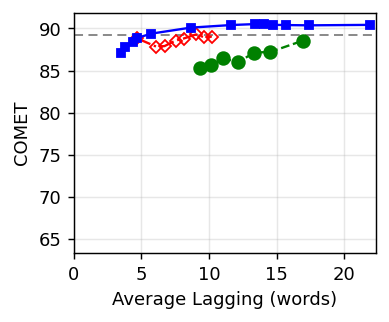}
\end{minipage}\hfill
\begin{minipage}{0.32\linewidth}
  \centering
  \includegraphics[width=\linewidth]{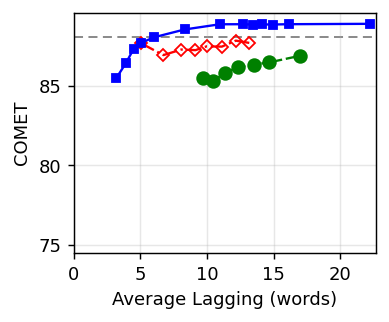}
\end{minipage}

\vspace{3pt}
\textbf{\footnotesize \code{continuation}}\\[1pt]
\begin{minipage}{0.32\linewidth}
  \centering
  \includegraphics[width=\linewidth]{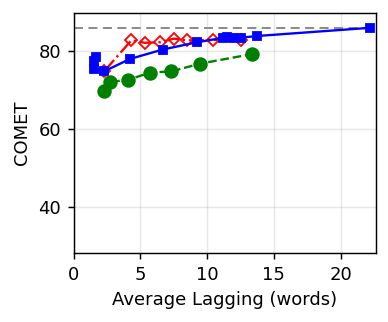}
\end{minipage}\hfill
\begin{minipage}{0.32\linewidth}
  \centering
  \includegraphics[width=\linewidth]{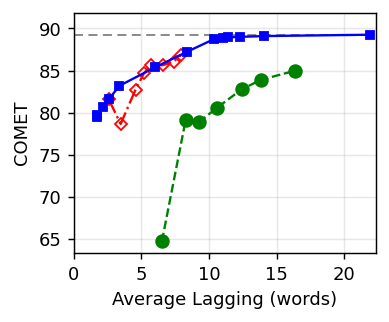}
\end{minipage}\hfill
\begin{minipage}{0.32\linewidth}
  \centering
  \includegraphics[width=\linewidth]{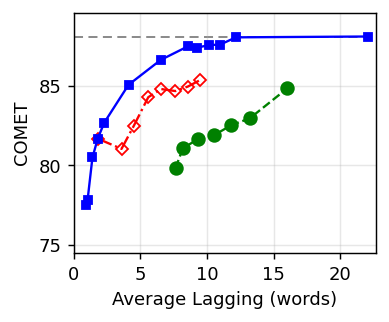}
\end{minipage}

\vspace{2pt}
\includegraphics[width=0.86\textwidth]{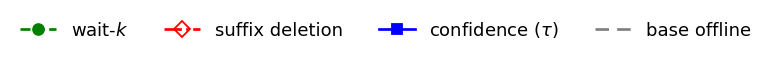}
\caption{The three latency mechanisms per model (rows) $\times$ EN$\rightarrow$\{DE,JA,ZH\} (columns) on FLEURS.}
\label{fig:knob-compare}
\end{figure*}

Table~\ref{tab:knob-compare} and Figure~\ref{fig:knob-compare} compare the three mechanisms for each model. 

\begin{table}[t]
\centering
% \footnotesize
\caption{Latency-mechanism AUC on FLEURS summarizing Figure~\ref{fig:knob-compare}: COMET over Average Lagging. \textbf{Bold}: best mechanism per row; $\star$: best per language.}
\label{tab:knob-compare}
\setlength{\tabcolsep}{6pt}
\begin{tabular}{@{}lrrr@{}}
\toprule
 & wait-$k$ & suffix del. & confidence \\
\midrule
% BEGIN KNOBCOMPARE_AUC
\multicolumn{4}{@{}l}{\code{base}} \\
EN$\rightarrow$DE & 47.9 & \textbf{75.1} & 75.0 \\
EN$\rightarrow$JA & 76.8 & 82.7 & \textbf{85.6} \\
EN$\rightarrow$ZH & 81.1 & 84.5 & \textbf{86.2} \\
\midrule
\multicolumn{4}{@{}l}{\code{prefix}} \\
EN$\rightarrow$DE & 65.0 & 73.7 & \textbf{78.9} \\
EN$\rightarrow$JA & 78.6 & 85.0 & \textbf{87.2}$^{\star}$ \\
EN$\rightarrow$ZH & 81.7 & 85.3 & \textbf{87.0}$^{\star}$ \\
\midrule
\multicolumn{4}{@{}l}{\code{continuation}} \\
EN$\rightarrow$DE & 74.8 & 80.2 & \textbf{81.9}$^{\star}$ \\
EN$\rightarrow$JA & 76.8 & 84.5 & \textbf{86.5} \\
EN$\rightarrow$ZH & 80.7 & 84.2 & \textbf{86.5} \\
\midrule
\multicolumn{4}{@{}l}{\code{full-control}} \\
EN$\rightarrow$DE & --- & --- & \textbf{76.0} \\
EN$\rightarrow$JA & --- & --- & \textbf{87.1} \\
EN$\rightarrow$ZH & --- & --- & \textbf{86.5} \\
% END KNOBCOMPARE_AUC
\bottomrule
\end{tabular}
\end{table}

The finetuned models achieve a much better quality--latency trade-off under any mechanism, though the confidence mechanism is clearly the best of the three in most cases.

Of the two finetuned variants, the \code{prefix} model is better than the \code{continuation} model, confirming the previous subsection's results. This can be explained by the fact that continuing from a previous translation is already implicit in the \code{prefix} model's training: the \code{continuation} task of explicitly teaching the model to continue a translation dilutes the useful supervision, adding little information while complicating the prefix/full-sentence data mix.

Wait-$k$'s rigid read-write schedule constrains what the model sees and forces it to commit on a fixed cadence. On the base model, this causes hallucinations, especially at low $k$, hence the low COMET score. The prefix-aware finetuned models reduce this degeneracy, but it remains uncompetitive.

Suffix deletion is not constrained by wait-$k$'s read schedule at low $k$: it can use the whole source prefix, so it translates better than wait-$k$ and occasionally beats the confidence mechanism. However, it can also hallucinate and repeat---content the deletion then discards, wasting computation---and it can only \emph{increase} latency by holding back the tail of the translation.

Overall, confidence is the best performing mechanism on all language pairs, models, and latency ranges (apart from \code{base} EN$\rightarrow$DE) but is particularly effective with the finetuned models. The mechanism is \emph{well-behaved at its endpoints}. At $\tau=0$ the rule reduces exactly to greedy decoding. At $\tau=1$ it waits for essentially the whole source before committing, recovering the offline translation. In between, the commit is \emph{confidence-adaptive} rather than positional. 
At negative $\tau$ it commits aggressively for lower latency, trading off translation quality.

\subsection{Calibration of commit/wait confidence}

When applied to the prefix or continuation models, which are trained to handle partial source inputs, the confidence mechanism becomes most effective. To distinguish the effect of stable-prefix supervision from generic full-sentence adaptation, this calibration analysis includes a full-sentence-only control for each language direction: a fresh adapter trained from the base model without stable-prefix examples. Its role is specifically to test whether ordinary full-sentence finetuning can account for changes in commit/wait confidence.

\begin{figure}[t]
\centering
\includegraphics[width=\linewidth]{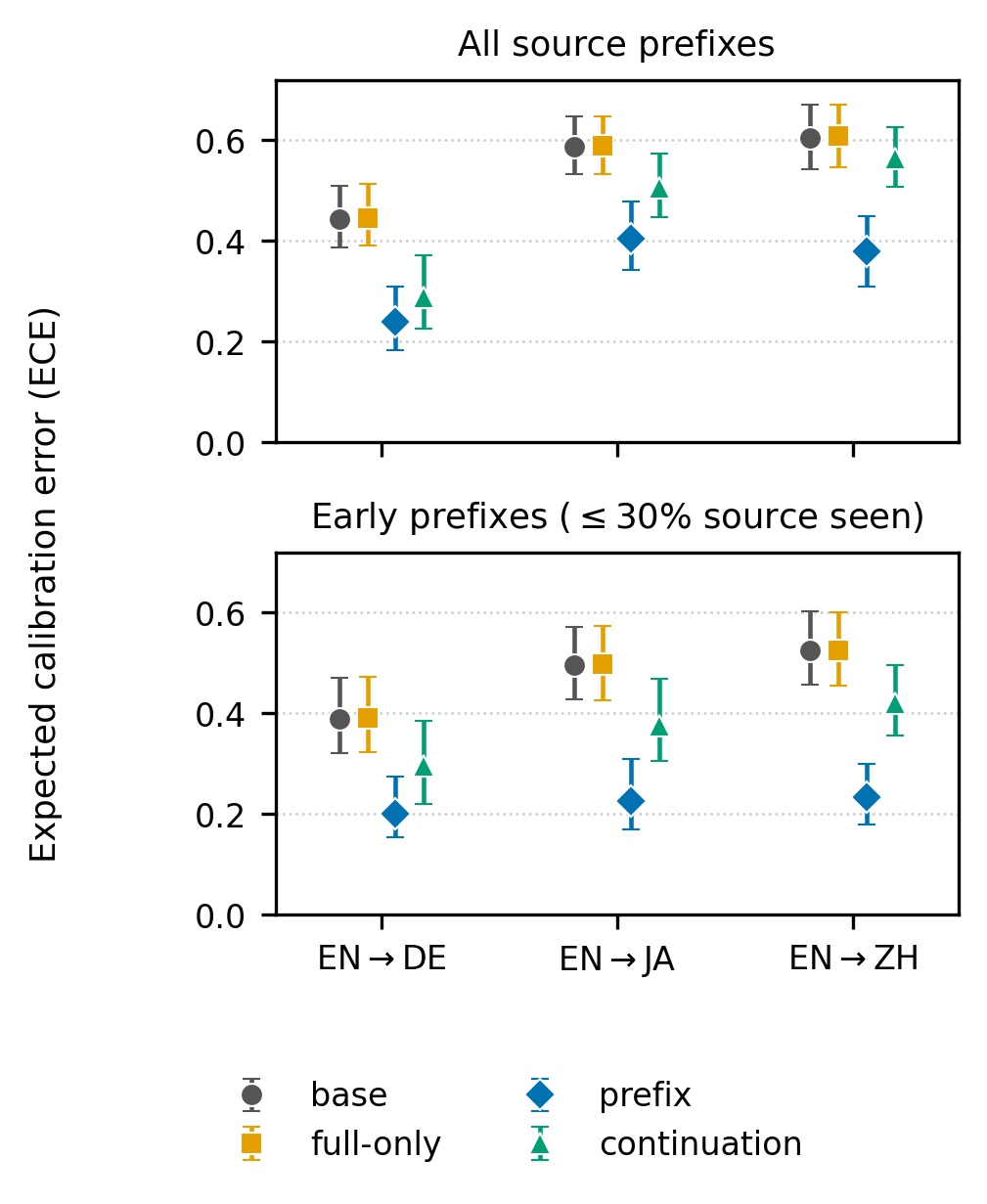}
\caption{Expected Calibration Error (lower is better) of normalized commit confidence $q_{\mathrm{go}}$ against the FLEURS stable-prefix oracle. Bars are $95\%$ intervals from 1,000 sentence-level bootstrap samples. \emph{Top:} all non-final source prefixes; \emph{bottom:} prefixes with at most $30\%$ of the source visible.}
\label{fig:commit-calibration}
\end{figure}

We test whether the continue-versus-stop signal is better calibrated using a FLEURS stable-prefix oracle (Figure~\ref{fig:commit-calibration}). For each of 64 test sentences, the base model translates the complete source and every proper source prefix; the oracle at each boundary is the running-max longest common prefix with the complete-source translation. We force-decode the previous oracle target, label each newly licensed target token as \textsc{go} followed by one \textsc{stop}, and score every model on these identical states using
\begin{equation}
q_{\mathrm{go}}=\frac{P(\text{best non-EOS})}{P(\text{best non-EOS})+P(\text{EOS})}.
\end{equation}

The full-sentence-only controls closely match \code{base} in all three directions, whereas the \code{prefix} model is substantially better calibrated: relative to \code{base}, ECE is reduced by $31$--$46\%$. On early prefixes with at most $30\%$ of the source visible, the reductions are $48$--$55\%$.

\code{continuation}'s ECE also improves on \code{base}, but less than \code{prefix}. The contrast with the full-sentence-only controls indicates that generic adaptation alone does not explain the calibration gain from stable-prefix training.

Training with the stable-prefix corpus thus not only produces models that are better at translating partial source inputs, but also improves commit/wait confidence calibration against the stable-prefix oracle.

\begin{figure*}[!t]\centering
\begin{minipage}{0.32\linewidth}
  \centering
\textbf{\footnotesize EN$\rightarrow$DE}\\[2pt]
  \includegraphics[width=\linewidth]{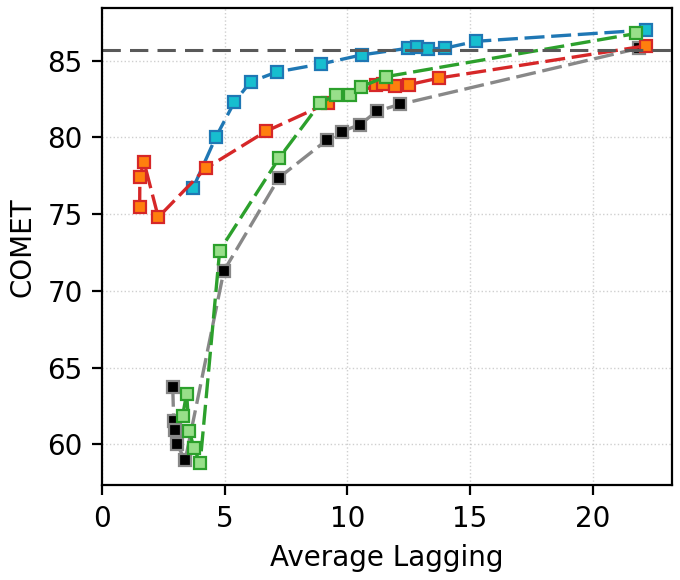}
\end{minipage}\hfill
\begin{minipage}{0.32\linewidth}
  \centering
\textbf{\footnotesize EN$\rightarrow$JA}\\[2pt]
  \includegraphics[width=\linewidth]{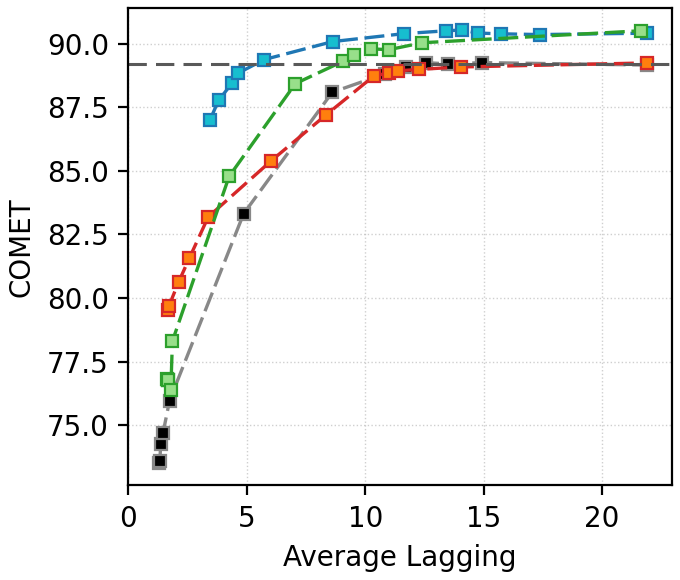}
\end{minipage}\hfill
\begin{minipage}{0.32\linewidth}
  \centering
\textbf{\footnotesize EN$\rightarrow$ZH}\\[2pt]
  \includegraphics[width=\linewidth]{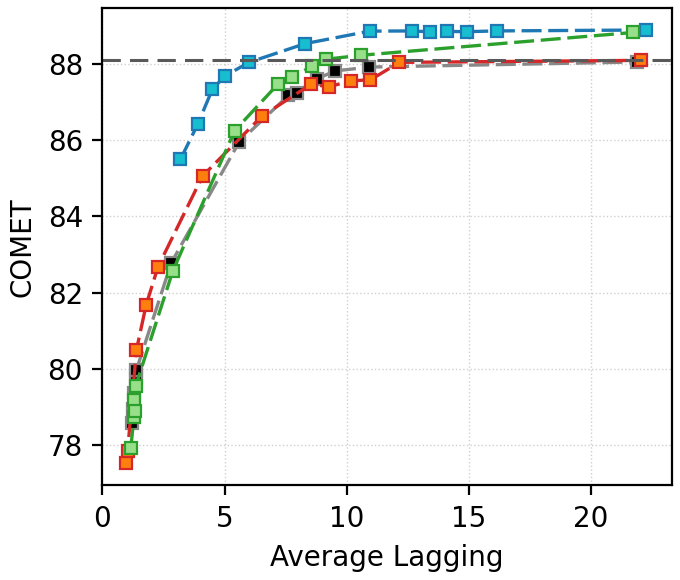}
\end{minipage}

\vspace{2pt}
\includegraphics[width=0.75\textwidth]{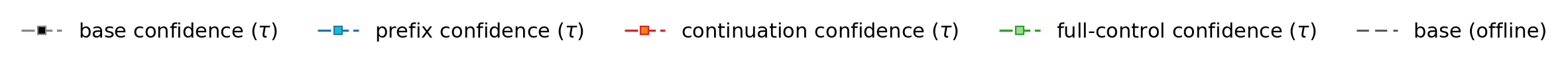}
\caption{Confidence quality--latency frontiers on FLEURS for \code{base}, \code{prefix}, \code{continuation}, and the \code{full-control} adapter.}
\label{fig:conf-fullcontrol}
\end{figure*}

The full-sentence-only model's full sentence translations (86.8/90.5/88.8) are better than the \code{base} model and similar to \code{prefix}, confirming the ``born-again'' effect. Its AUC (last three rows of Table~\ref{tab:knob-compare}) likewise rises above \code{base} and nearly matches \code{prefix} because its full sentence translation pulls up the entire curve. However, as Figure~\ref{fig:conf-fullcontrol} shows, its latency-quality tradeoff consistently drops relative to \code{prefix} when pushed for lower latency, reflecting its uncalibrated performance.

\subsection{Evaluations on other test sets}

\begin{figure*}[!t]
\centering
\begin{minipage}{0.32\linewidth}
  \centering
\textbf{\footnotesize EN$\rightarrow$DE}
\end{minipage}\hfill
\begin{minipage}{0.32\linewidth}
  \centering
\textbf{\footnotesize EN$\rightarrow$JA}
\end{minipage}\hfill
\begin{minipage}{0.32\linewidth}
  \centering
\textbf{\footnotesize EN$\rightarrow$ZH}
\end{minipage}

\vspace{2pt}
\textbf{FLEURS}\\[1pt]
\begin{minipage}{0.32\linewidth}
  \centering
  \includegraphics[width=\linewidth]{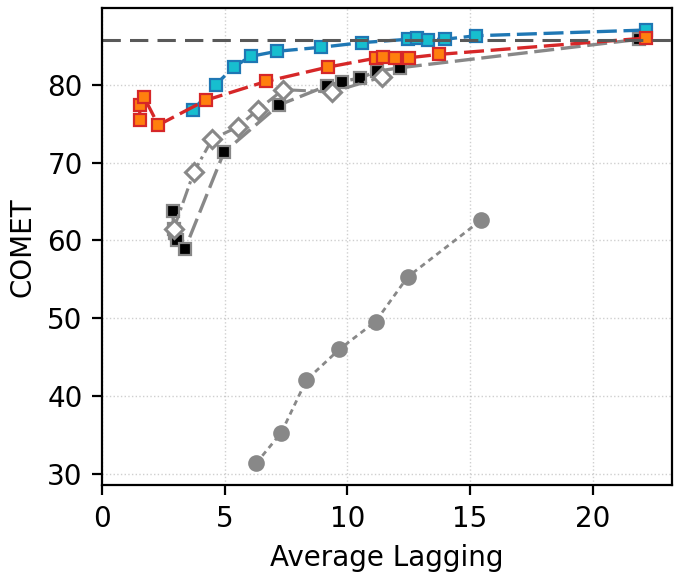}
\end{minipage}\hfill
\begin{minipage}{0.32\linewidth}
  \centering
  \includegraphics[width=\linewidth]{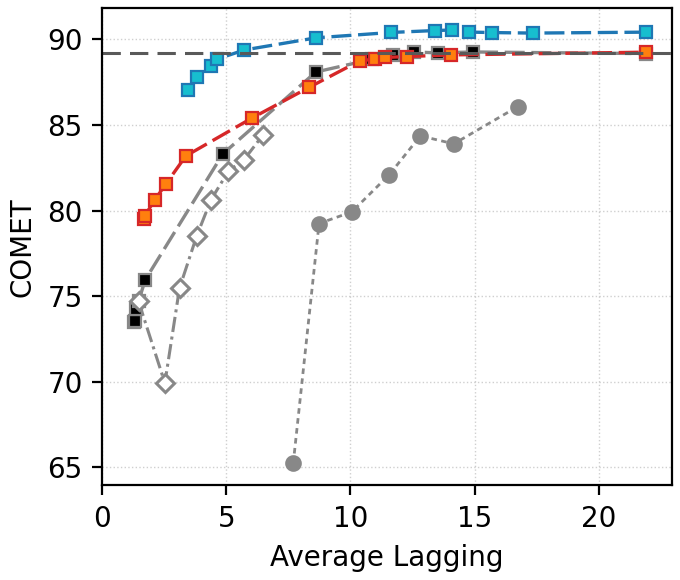}
\end{minipage}\hfill
\begin{minipage}{0.32\linewidth}
  \centering
  \includegraphics[width=\linewidth]{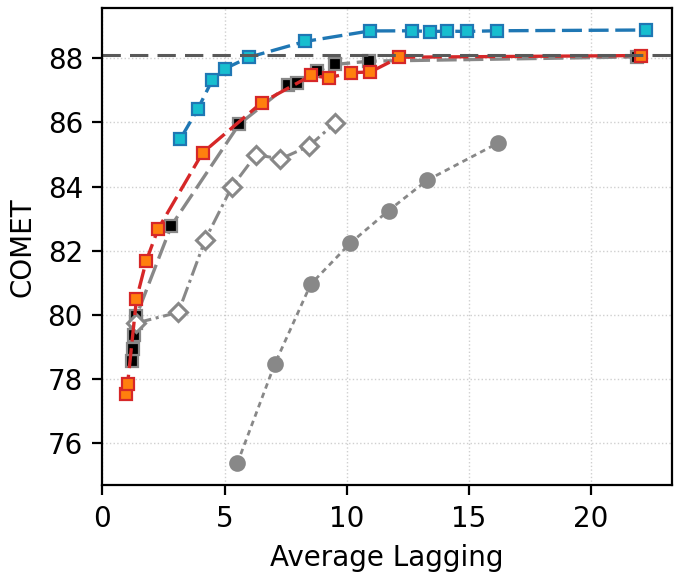}
\end{minipage}

\vspace{3pt}
\textbf{WMT24{+}{+}}\\[1pt]
\begin{minipage}{0.32\linewidth}
  \centering
  \includegraphics[width=\linewidth]{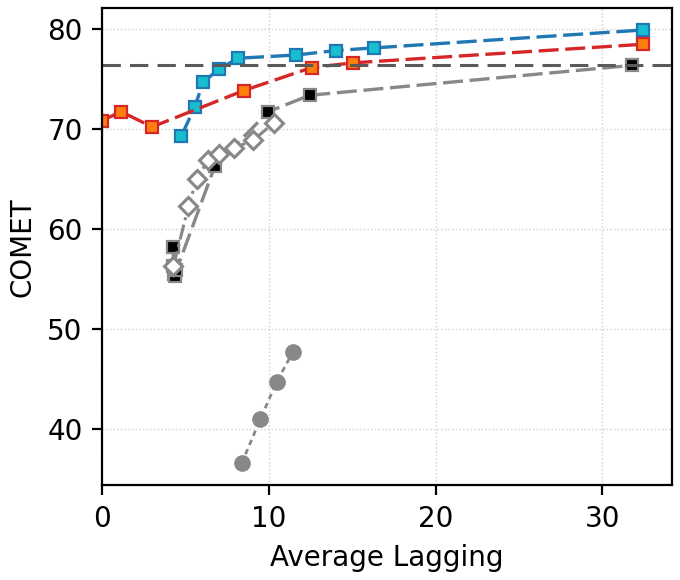}
\end{minipage}\hfill
\begin{minipage}{0.32\linewidth}
  \centering
  \includegraphics[width=\linewidth]{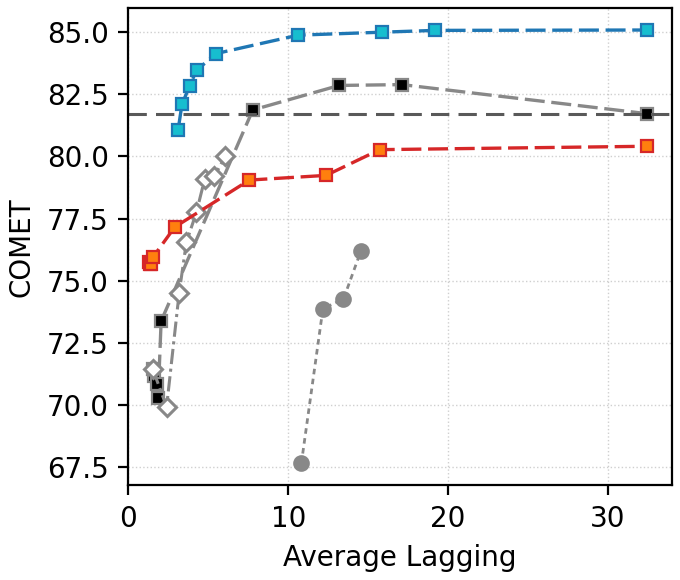}
\end{minipage}\hfill
\begin{minipage}{0.32\linewidth}
  \centering
  \includegraphics[width=\linewidth]{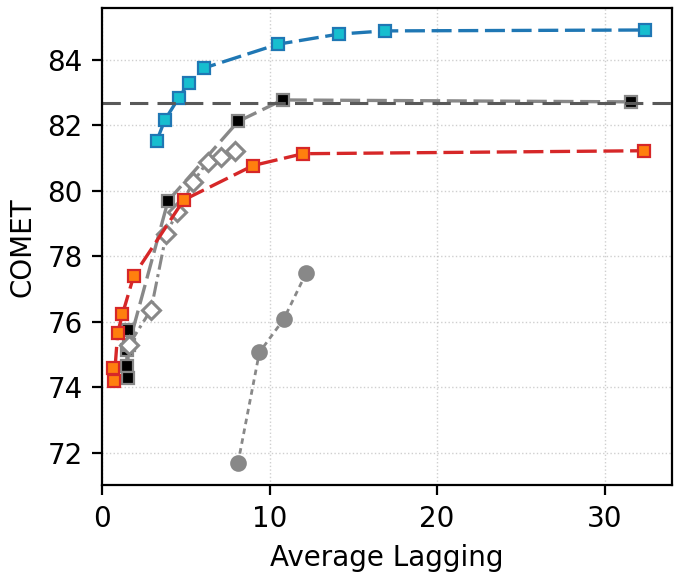}
\end{minipage}

\vspace{3pt}
\textbf{CoVoST2}\\[1pt]
\begin{minipage}{0.32\linewidth}
  \centering
  \includegraphics[width=\linewidth]{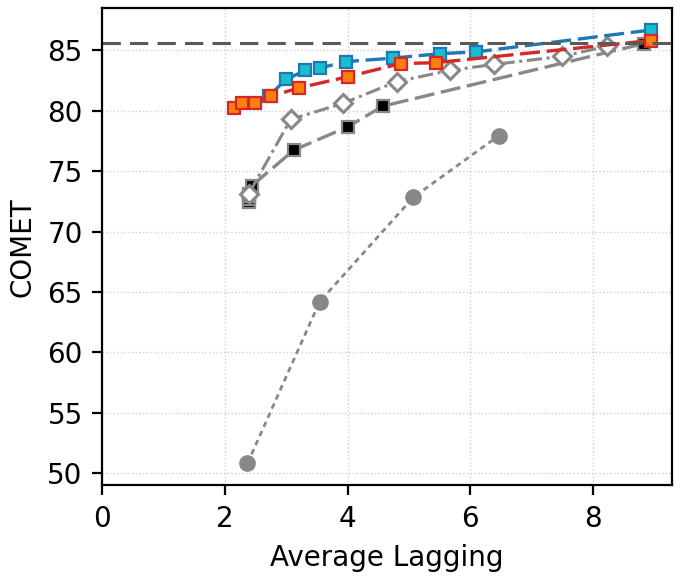}
\end{minipage}\hfill
\begin{minipage}{0.32\linewidth}
  \centering
  \includegraphics[width=\linewidth]{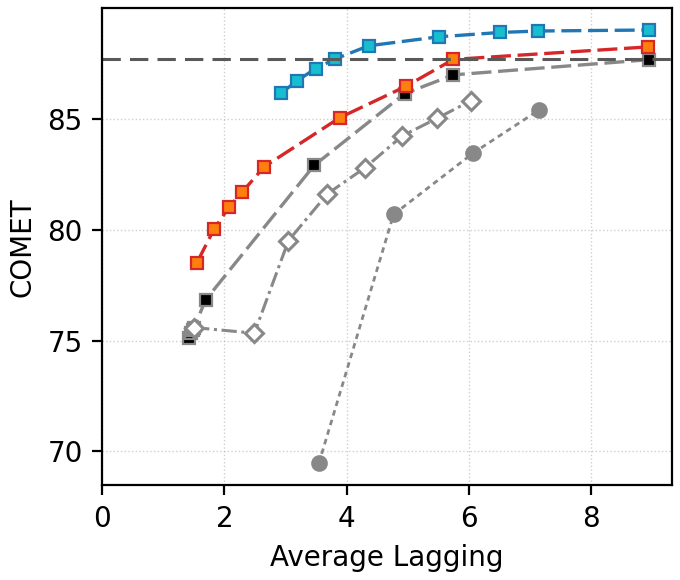}
\end{minipage}\hfill
\begin{minipage}{0.32\linewidth}
  \centering
  \includegraphics[width=\linewidth]{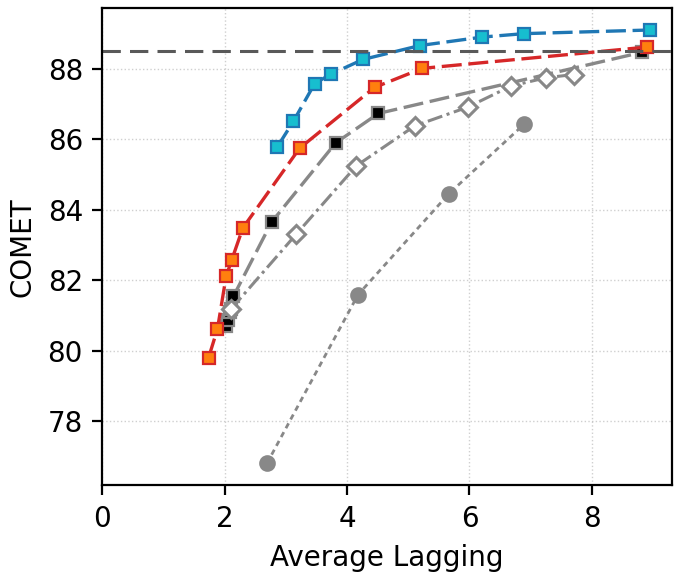}
\end{minipage}

\vspace{2pt}
\includegraphics[width=0.9\textwidth]{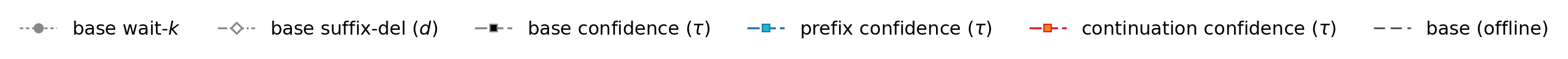}
\caption{COMET quality--latency tradeoff on FLEURS, WMT24++, and CoVoST~2 (rows) $\times$ EN$\rightarrow$\{DE,JA,ZH\} (columns). Each panel compares \code{base} under wait-$k$, suffix deletion, and confidence with \code{prefix} and \code{continuation} under confidence.}
\label{fig:sched-robust}
\end{figure*}

The FLEURS comparison in the previous section carries over to the other test sets. Table~\ref{tab:sched-robust-auc} and Figure~\ref{fig:sched-robust} repeat the evaluation on WMT24++ and CoVoST2 alongside FLEURS: the \code{base} model is swept with wait-$k$, suffix deletion, and confidence, while the finetuned models use the confidence mechanism.

The explicit \code{base} confidence sweep controls for the inference mechanism when comparing the pre-trained and finetuned models. By COMET AUC, \code{prefix} with confidence improves over \code{base} with confidence in all nine dataset--language panels. \code{prefix} has the higher quality ceiling, while \code{continuation} is useful for the lowest-latency frontier.

Appendix Table~\ref{tab:sched-robust-auc-metricx} and Figure~\ref{fig:sched-robust-metricx} present the corresponding MetricX results; they support the same conclusions as the COMET-based analysis.

\begin{table}[t]
\centering
\footnotesize
\caption{COMET quality--latency frontier AUC summarizing Figure~\ref{fig:sched-robust} over Average Lagging, for EN$\rightarrow$\{DE,JA,ZH\}. Each column scores one sweep on the shared latency interval for that dataset--language panel; best per row in \textbf{bold}.}
\label{tab:sched-robust-auc}
\setlength{\tabcolsep}{2pt}
\begin{tabular}{@{}lrrrrr@{}}
\toprule
 & \multicolumn{3}{c}{\code{base}} & \multicolumn{2}{c}{ft.\,+\,conf.} \\
\cmidrule(lr){2-4}\cmidrule(l){5-6}
 & wait-$k$ & suf.\,del. & conf. & \code{prefix} & \code{cont.} \\
\midrule
% BEGIN SCHED_ROBUST_AUC
\multicolumn{6}{@{}l}{\textbf{FLEURS}} \\
DE & 48.2 & 75.1 & 75.1 & 79.0 & \textbf{81.9} \\
JA & 77.0 & 82.7 & 85.6 & \textbf{87.3} & 86.5 \\
ZH & 81.1 & 84.5 & 86.2 & \textbf{87.0} & 86.5 \\
\midrule
\multicolumn{6}{@{}l}{\textbf{WMT24++}} \\
DE & 43.7 & 63.6 & 65.1 & 69.6 & \textbf{74.6} \\
JA & 73.1 & 79.2 & 80.6 & \textbf{83.7} & 79.3 \\
ZH & 75.7 & 80.4 & 81.4 & \textbf{83.3} & 80.3 \\
\midrule
\multicolumn{6}{@{}l}{\textbf{CoVoST2}} \\
DE & 68.2 & 80.5 & 78.0 & 81.3 & \textbf{83.1} \\
JA & 77.2 & 82.1 & 83.3 & 84.4 & \textbf{85.0} \\
ZH & 81.8 & 85.0 & 85.2 & \textbf{86.5} & \textbf{86.5} \\
% END SCHED_ROBUST_AUC
\bottomrule
\end{tabular}
\end{table}

%%%%%%%%%%%%%%%%%%%%%%%%%%%%%%%%%%%%%%%%%%%%%%%%%%
\section{Conclusion}

We presented a method for identifying \emph{stable training prefixes}, which are translations of source-sentence prefixes that are in common with their corresponding full-sentence translations.
Rather than translating the currently available source prefix in its entirety or applying a fixed truncation rule, the model learns how much target content to commit. 
% We derive this supervision from sentence-level stable prefixes, mix it with full-sentence translations, and prompt the model with a single force-decode turn that carries the committed target forward as more source arrives.
This provides a dynamic, context-sensitive approach to determining how much of an input prefix to translate, in the context of simultaneous translation.
% way to make a sentence-trained, decoder-only MT model prefix-aware when its source read schedule is imposed by an upstream system. 
In our experiments, we mixed this data with normal, full-sentence training data and fine-tuned a large language model.
Inference with the model focused on a prefix-decoding strategy, a streaming approach that does not modify the output, once committed, and is therefore flicker-free by design.
Experiments with Qwen3-8B on EN$\rightarrow$DE, EN$\rightarrow$JA, and EN$\rightarrow$ZH show improved latency-accuracy trade-offs under a number of decoding strategies.

We trained two model variants, \code{prefix} and \code{continuation}, to handle initial and ongoing translation tasks.
Of these, \code{prefix} dominates most of the Pareto frontier in its latency range. The \code{continuation} model can reach slightly lower latency points, but with a trade-off in translation quality.
Although the \code{prefix} model used only initial-prefix source segmentations, its performance generalized to random read schedules and continuations, a result that held across language pairs, datasets, and evaluation metrics.

The training also substantially improves calibration of commit/wait confidence against a FLEURS stable-prefix oracle, reducing ECE by $31$--$46\%$ overall and $48$--$55\%$ on early source prefixes. 
A single confidence threshold is the best of the three training-free latency mechanisms on the finetuned models we tested. 
It outperforms wait-$k$ and suffix-deletion in all language pairs as measured by AUC, and extends the reachable frontier to higher and lower latency.

%%%%%%%%%%%%%%%%%%%%%%%%%%%%%%%%%%%%%%%%%%%%%%%%%%
\section{Limitations}

Our experiments cover only three language pairs, all originating in English, and we fine-tune under only a single translation engine; further experimentation is required to test whether the results generalize.

% \begin{itemize}
% \item Results cover three language pairs (EN$\rightarrow$DE, EN$\rightarrow$JA, EN$\rightarrow$ZH), and one base model (Qwen3-8B); transfer to further pairs and model families is untested.
% \item Evaluation uses textual prefixes of reference transcripts rather than hypotheses from a streaming ASR system. We therefore do not measure the effects of ASR errors, hypothesis revisions, or ASR timing on translation.
% \item We report COMET, MetricX, Average Lagging. We do not use other latency measures (DAL, LAAL, AP, etc.) in this work.
% \item Because committed output is monotonic---force-decoding freezes the committed prefix and no operating mode revises it---the revision (flicker) rate is exactly zero by construction, so we do not report a direct revision-rate measurement.
% \item We only use greedy decoding; no sampling/temperature study.
% \item Calibration is measured on 64 FLEURS test sentences per direction against a base-derived stable-prefix oracle. It concerns the model's commit/wait confidence, not generic language-model calibration or human judgments of semantic support. The original prefix adapters do not preserve sufficient training-corpus provenance to guarantee that the full-sentence-only control is exactly matched in data exposure and stopping time.
% \item Qwen3-8B was released in April 2025, so its pretraining data may overlap with our validation or test sets. We did not audit for such contamination.
% \end{itemize}

\bibliography{references}

@inproceedings{ma-etal-2019-stacl,
  address =       {Florence, Italy},
  author =        {Ma, Mingbo and Huang, Liang and Xiong, Hao and
                   Zheng, Renjie and Liu, Kaibo and Zheng, Baigong and
                   Zhang, Chuanqiang and He, Zhongjun and Liu, Hairong and
                   Li, Xing and Wu, Hua and Wang, Haifeng},
  booktitle =     {Proceedings of the 57th Annual Meeting of the
                   Association for Computational Linguistics},
  editor =        {Korhonen, Anna and Traum, David and
                   M{\`a}rquez, Llu{\'i}s},
  month =         jul,
  pages =         {3025--3036},
  publisher =     {Association for Computational Linguistics},
  title =         {{STACL}: Simultaneous Translation with Implicit
                   Anticipation and Controllable Latency using
                   Prefix-to-Prefix Framework},
  year =          {2019},
  doi =           {10.18653/v1/P19-1289},
  url =           {https://aclanthology.org/P19-1289/},
}

@inproceedings{gu-etal-2017-learning,
  address =       {Valencia, Spain},
  author =        {Gu, Jiatao and Neubig, Graham and Cho, Kyunghyun and
                   Li, Victor O.K.},
  booktitle =     {Proceedings of the 15th Conference of the {E}uropean
                   Chapter of the Association for Computational
                   Linguistics: Volume 1, Long Papers},
  editor =        {Lapata, Mirella and Blunsom, Phil and
                   Koller, Alexander},
  month =         apr,
  pages =         {1053--1062},
  publisher =     {Association for Computational Linguistics},
  title =         {Learning to Translate in Real-time with Neural
                   Machine Translation},
  year =          {2017},
  url =           {https://aclanthology.org/E17-1099/},
}

@inproceedings{zheng-etal-2020-simultaneous,
  address =       {Online},
  author =        {Zheng, Baigong and Liu, Kaibo and Zheng, Renjie and
                   Ma, Mingbo and Liu, Hairong and Huang, Liang},
  booktitle =     {Proceedings of the 58th Annual Meeting of the
                   Association for Computational Linguistics},
  editor =        {Jurafsky, Dan and Chai, Joyce and Schluter, Natalie and
                   Tetreault, Joel},
  month =         jul,
  pages =         {2847--2853},
  publisher =     {Association for Computational Linguistics},
  title =         {Simultaneous Translation Policies: From Fixed to
                   Adaptive},
  year =          {2020},
  doi =           {10.18653/v1/2020.acl-main.254},
  url =           {https://aclanthology.org/2020.acl-main.254/},
}

@inproceedings{arivazhagan-etal-2019-retranslation,
  author =        {Arivazhagan, Naveen and Cherry, Colin and Te, I and
                   Macherey, Wolfgang and Baljekar, Pallavi and
                   Foster, George},
  booktitle =     {2020 IEEE International Conference on Acoustics,
                   Speech and Signal Processing (ICASSP)},
  pages =         {7919--7923},
  title =         {Re-Translation Strategies for Long Form,
                   Simultaneous, Spoken Language Translation},
  year =          {2020},
  doi =           {10.1109/ICASSP40776.2020.9054585},
  url =           {https://arxiv.org/abs/1912.03393},
}

@inproceedings{polak-etal-2022-cuni,
  address =       {Dublin, Ireland (in-person and online)},
  author =        {Pol{\'a}k, Peter and Pham, Ngoc-Quan and
                   Nguyen, Tuan Nam and Liu, Danni and Mullov, Carlos and
                   Niehues, Jan and Bojar, Ond{\v{r}}ej and
                   Waibel, Alexander},
  booktitle =     {Proceedings of the 19th International Conference on
                   Spoken Language Translation (IWSLT 2022)},
  editor =        {Salesky, Elizabeth and Federico, Marcello and
                   Costa-juss{\`a}, Marta},
  month =         may,
  pages =         {277--285},
  publisher =     {Association for Computational Linguistics},
  title =         {{CUNI}-{KIT} System for Simultaneous Speech
                   Translation Task at {IWSLT} 2022},
  year =          {2022},
  doi =           {10.18653/v1/2022.iwslt-1.24},
  url =           {https://aclanthology.org/2022.iwslt-1.24/},
}

@inproceedings{bentivogli-etal-2021-cascade,
  address =       {Online},
  author =        {Bentivogli, Luisa and Cettolo, Mauro and Gaido, Marco and
                   Karakanta, Alina and Martinelli, Alberto and
                   Negri, Matteo and Turchi, Marco},
  booktitle =     {Proceedings of the 59th Annual Meeting of the
                   Association for Computational Linguistics and the
                   11th International Joint Conference on Natural
                   Language Processing (Volume 1: Long Papers)},
  editor =        {Zong, Chengqing and Xia, Fei and Li, Wenjie and
                   Navigli, Roberto},
  month =         aug,
  pages =         {2873--2887},
  publisher =     {Association for Computational Linguistics},
  title =         {Cascade versus Direct Speech Translation: Do the
                   Differences Still Make a Difference?},
  year =          {2021},
  doi =           {10.18653/v1/2021.acl-long.224},
  url =           {https://aclanthology.org/2021.acl-long.224/},
}

@misc{conneau-etal-2022-fleurs,
  author =        {Conneau, Alexis and Ma, Min and Khanuja, Simran and
                   Zhang, Yu and Axelrod, Vera and Dalmia, Siddharth and
                   Riesa, Jason and Rivera, Clara and Bapna, Ankur},
  title =         {{FLEURS}: Few-shot Learning Evaluation of Universal
                   Representations of Speech},
  year =          {2022},
  doi =           {10.48550/arXiv.2205.12446},
  url =           {https://arxiv.org/abs/2205.12446},
}

@inproceedings{deutsch-etal-2025-wmt24,
  address =       {Vienna, Austria},
  author =        {Deutsch, Daniel and Briakou, Eleftheria and
                   Caswell, Isaac Rayburn and Finkelstein, Mara and
                   Galor, Rebecca and Juraska, Juraj and Kovacs, Geza and
                   Lui, Alison and Rei, Ricardo and Riesa, Jason and
                   Rijhwani, Shruti and Riley, Parker and
                   Salesky, Elizabeth and Trabelsi, Firas and
                   Winkler, Stephanie and Zhang, Biao and
                   Freitag, Markus},
  booktitle =     {Findings of the Association for Computational
                   Linguistics: ACL 2025},
  editor =        {Che, Wanxiang and Nabende, Joyce and
                   Shutova, Ekaterina and Pilehvar, Mohammad Taher},
  month =         jul,
  pages =         {12257--12284},
  publisher =     {Association for Computational Linguistics},
  title =         {{WMT}24++: Expanding the Language Coverage of {WMT}24
                   to 55 Languages {\&} Dialects},
  year =          {2025},
  doi =           {10.18653/v1/2025.findings-acl.634},
  isbn =          {979-8-89176-256-5},
  url =           {https://aclanthology.org/2025.findings-acl.634/},
}

@misc{wang-etal-2020-covost2,
  author =        {Wang, Changhan and Wu, Anne and Pino, Juan},
  title =         {{CoVoST} 2 and Massively Multilingual Speech-to-Text
                   Translation},
  year =          {2020},
  doi =           {10.48550/arXiv.2007.10310},
  url =           {https://arxiv.org/abs/2007.10310},
}

@article{cho-esipova-2016-can,
  author =        {Kyunghyun Cho and Masha Esipova},
  journal =       {CoRR},
  title =         {Can neural machine translation do simultaneous
                   translation?},
  volume =        {abs/1606.02012},
  year =          {2016},
  bibsource =     {dblp computer science bibliography, https://dblp.org},
  url =           {http://arxiv.org/abs/1606.02012},
}

@inproceedings{arivazhagan-etal-2020-translation,
  address =       {Online},
  author =        {Arivazhagan, Naveen and Cherry, Colin and
                   Macherey, Wolfgang and Foster, George},
  booktitle =     {Proceedings of the 17th International Conference on
                   Spoken Language Translation},
  editor =        {Federico, Marcello and Waibel, Alex and Knight, Kevin and
                   Nakamura, Satoshi and Ney, Hermann and Niehues, Jan and
                   St{\"u}ker, Sebastian and Wu, Dekai and
                   Mariani, Joseph and Yvon, Francois},
  month =         jul,
  pages =         {220--227},
  publisher =     {Association for Computational Linguistics},
  title =         {Re-translation versus Streaming for Simultaneous
                   Translation},
  year =          {2020},
  doi =           {10.18653/v1/2020.iwslt-1.27},
  url =           {https://aclanthology.org/2020.iwslt-1.27/},
}

@inproceedings{huang-etal-2020-simultaneous,
  address =       {Online},
  author =        {Huang, Liang and Cherry, Colin and Ma, Mingbo and
                   Arivazhagan, Naveen and He, Zhongjun},
  booktitle =     {Proceedings of the 2020 Conference on Empirical
                   Methods in Natural Language Processing: Tutorial
                   Abstracts},
  editor =        {Villavicencio, Aline and Van Durme, Benjamin},
  month =         nov,
  pages =         {34--36},
  publisher =     {Association for Computational Linguistics},
  title =         {Simultaneous Translation},
  year =          {2020},
  doi =           {10.18653/v1/2020.emnlp-tutorials.6},
  url =           {https://aclanthology.org/2020.emnlp-tutorials.6/},
}

@inproceedings{yang-etal-2025-large-language,
  address =       {Vienna, Austria},
  author =        {Yang, Zhengdong and Shimizu, Shuichiro and Yu, Yahan and
                   Chu, Chenhui},
  booktitle =     {Findings of the Association for Computational
                   Linguistics: ACL 2025},
  editor =        {Che, Wanxiang and Nabende, Joyce and
                   Shutova, Ekaterina and Pilehvar, Mohammad Taher},
  month =         jul,
  pages =         {20298--20315},
  publisher =     {Association for Computational Linguistics},
  title =         {When Large Language Models Meet Speech: A Survey on
                   Integration Approaches},
  year =          {2025},
  doi =           {10.18653/v1/2025.findings-acl.1041},
  isbn =          {979-8-89176-256-5},
  url =           {https://aclanthology.org/2025.findings-acl.1041/},
}

@inproceedings{ouyang-etal-2025-infinisst,
  address =       {Vienna, Austria},
  author =        {Ouyang, Siqi and Xu, Xi and Li, Lei},
  booktitle =     {Findings of the Association for Computational
                   Linguistics: ACL 2025},
  editor =        {Che, Wanxiang and Nabende, Joyce and
                   Shutova, Ekaterina and Pilehvar, Mohammad Taher},
  month =         jul,
  pages =         {3032--3046},
  publisher =     {Association for Computational Linguistics},
  title =         {{I}nfini{SST}: Simultaneous Translation of Unbounded
                   Speech with Large Language Model},
  year =          {2025},
  doi =           {10.18653/v1/2025.findings-acl.157},
  isbn =          {979-8-89176-256-5},
  url =           {https://aclanthology.org/2025.findings-acl.157/},
}

@inproceedings{wang-etal-2025-conversational,
  address =       {Vienna, Austria (in-person and online)},
  author =        {Wang, Minghan and Vu, Thuy-Trang and Wang, Yuxia and
                   Shareghi, Ehsan and Haffari, Gholamreza},
  booktitle =     {Proceedings of the 22nd International Conference on
                   Spoken Language Translation (IWSLT 2025)},
  editor =        {Salesky, Elizabeth and Federico, Marcello and
                   Anastasopoulos, Antonis},
  month =         jul,
  pages =         {93--105},
  publisher =     {Association for Computational Linguistics},
  title =         {Conversational {S}imul{MT}: Efficient Simultaneous
                   Translation with Large Language Models},
  year =          {2025},
  doi =           {10.18653/v1/2025.iwslt-1.8},
  isbn =          {979-8-89176-272-5},
  url =           {https://aclanthology.org/2025.iwslt-1.8/},
}

@inproceedings{agostinelli-etal-2024-simul,
  address =       {Bangkok, Thailand},
  author =        {Agostinelli, Victor and Wild, Max and Raffel, Matthew and
                   Fuad, Kazi and Chen, Lizhong},
  booktitle =     {Proceedings of the 62nd Annual Meeting of the
                   Association for Computational Linguistics (Volume 1:
                   Long Papers)},
  editor =        {Ku, Lun-Wei and Martins, Andre and Srikumar, Vivek},
  month =         aug,
  pages =         {10530--10541},
  publisher =     {Association for Computational Linguistics},
  title =         {Simul-{LLM}: A Framework for Exploring High-Quality
                   Simultaneous Translation with Large Language Models},
  year =          {2024},
  doi =           {10.18653/v1/2024.acl-long.567},
  url =           {https://aclanthology.org/2024.acl-long.567/},
}

@inproceedings{fu-etal-2025-llms,
  address =       {Vienna, Austria},
  author =        {Fu, Biao and Liao, Minpeng and Fan, Kai and
                   Li, Chengxi and Zhang, Liang and Chen, Yidong and
                   Shi, Xiaodong},
  booktitle =     {Findings of the Association for Computational
                   Linguistics: ACL 2025},
  editor =        {Che, Wanxiang and Nabende, Joyce and
                   Shutova, Ekaterina and Pilehvar, Mohammad Taher},
  month =         jul,
  pages =         {20372--20395},
  publisher =     {Association for Computational Linguistics},
  title =         {{LLM}s Can Achieve High-quality Simultaneous Machine
                   Translation as Efficiently as Offline},
  year =          {2025},
  doi =           {10.18653/v1/2025.findings-acl.1045},
  isbn =          {979-8-89176-256-5},
  url =           {https://aclanthology.org/2025.findings-acl.1045/},
}

@misc{qwen2024,
  author =        {Yang, An and Li, Anfeng and Yang, Baosong and
                   Zhang, Beichen and Hui, Binyuan and Zheng, Bo and
                   Yu, Bowen and Gao, Chang and Huang, Chengen and
                   Lv, Chenxu and Zheng, Chujie and Liu, Dayiheng and
                   Zhou, Fan and Huang, Fei and Hu, Feng and Ge, Hao and
                   Wei, Haoran and Lin, Huan and Tang, Jialong and
                   Yang, Jian and Tu, Jianhong and Zhang, Jianwei and
                   Yang, Jianxin and Yang, Jiaxi and Zhou, Jing and
                   Zhou, Jingren and Lin, Junyang and Dang, Kai and
                   Bao, Keqin and Yang, Kexin and Yu, Le and
                   Deng, Lianghao and Li, Mei and Xue, Mingfeng and
                   Li, Mingze and Zhang, Pei and Wang, Peng and Zhu, Qin and
                   Men, Rui and Gao, Ruize and Liu, Shixuan and
                   Luo, Shuang and Li, Tianhao and Tang, Tianyi and
                   Yin, Wenbiao and Ren, Xingzhang and Wang, Xinyu and
                   Zhang, Xinyu and Ren, Xuancheng and Fan, Yang and
                   Su, Yang and Zhang, Yichang and Zhang, Yinger and
                   Wan, Yu and Liu, Yuqiong and Wang, Zekun and
                   Cui, Zeyu and Zhang, Zhenru and Zhou, Zhipeng and
                   Qiu, Zihan},
  title =         {{Qwen3} Technical Report},
  year =          {2025},
  doi =           {10.48550/arXiv.2505.09388},
  url =           {https://arxiv.org/abs/2505.09388},
}

@inproceedings{niehues-etal-2016-dynamic,
  author =        {Jan Niehues and Thai Son Nguyen and Eunah Cho and
                   Thanh-Le Ha and Kevin Kilgour and Markus Müller and
                   Matthias Sperber and Sebastian Stüker and
                   Alex Waibel},
  booktitle =     {{Interspeech 2016}},
  pages =         {2513--2517},
  title =         {{Dynamic Transcription for Low-Latency Speech
                   Translation}},
  year =          {2016},
  doi =           {10.21437/Interspeech.2016-154},
  issn =          {2958-1796},
}

@inproceedings{sen-etal-2023-self,
  address =       {Dubrovnik, Croatia},
  author =        {Sen, Sukanta and Sennrich, Rico and Zhang, Biao and
                   Haddow, Barry},
  booktitle =     {Proceedings of the 17th Conference of the European
                   Chapter of the Association for Computational
                   Linguistics},
  editor =        {Vlachos, Andreas and Augenstein, Isabelle},
  month =         may,
  pages =         {3734--3744},
  publisher =     {Association for Computational Linguistics},
  title =         {Self-training Reduces Flicker in Retranslation-based
                   Simultaneous Translation},
  year =          {2023},
  doi =           {10.18653/v1/2023.eacl-main.270},
  url =           {https://aclanthology.org/2023.eacl-main.270/},
}

@inproceedings{liu-etal-2020-low,
  author =        {Danni Liu and Gerasimos Spanakis and Jan Niehues},
  booktitle =     {Interspeech 2020},
  pages =         {3620--3624},
  title =         {Low-Latency Sequence-to-Sequence Speech Recognition
                   and Translation by Partial Hypothesis Selection},
  year =          {2020},
  doi =           {10.21437/Interspeech.2020-2897},
  issn =          {2958-1796},
}

@inproceedings{niehues-etal-2018-low,
  author =        {Jan Niehues and Ngoc-Quan Pham and Thanh-Le Ha and
                   Matthias Sperber and Alex Waibel},
  booktitle =     {{Interspeech 2018}},
  pages =         {1293--1297},
  title =         {{Low-Latency Neural Speech Translation}},
  year =          {2018},
  doi =           {10.21437/Interspeech.2018-1055},
}

@inproceedings{kocmi-etal-2024-findings,
  address =       {Miami, Florida, USA},
  author =        {Kocmi, Tom and Avramidis, Eleftherios and
                   Bawden, Rachel and Bojar, Ond{\v{r}}ej and
                   Dvorkovich, Anton and Federmann, Christian and
                   Fishel, Mark and Freitag, Markus and Gowda, Thamme and
                   Grundkiewicz, Roman and Haddow, Barry and
                   Karpinska, Marzena and Koehn, Philipp and
                   Marie, Benjamin and Monz, Christof and Murray, Kenton and
                   Nagata, Masaaki and Popel, Martin and
                   Popovi{\'c}, Maja and Shmatova, Mariya and
                   Steingr{\'i}msson, Steinth{\'o}r and
                   Zouhar, Vil{\'e}m},
  booktitle =     {Proceedings of the Ninth Conference on Machine
                   Translation},
  editor =        {Haddow, Barry and Kocmi, Tom and Koehn, Philipp and
                   Monz, Christof},
  month =         nov,
  pages =         {1--46},
  publisher =     {Association for Computational Linguistics},
  title =         {Findings of the {WMT}24 General Machine Translation
                   Shared Task: The {LLM} Era Is Here but {MT} Is Not
                   Solved Yet},
  year =          {2024},
  doi =           {10.18653/v1/2024.wmt-1.1},
  url =           {https://aclanthology.org/2024.wmt-1.1/},
}

@inproceedings{kocmi-etal-2025-findings,
  address =       {Suzhou, China},
  author =        {Kocmi, Tom and Artemova, Ekaterina and
                   Avramidis, Eleftherios and Bawden, Rachel and
                   Bojar, Ond{\v{r}}ej and Dranch, Konstantin and
                   Dvorkovich, Anton and Dukanov, Sergey and
                   Fishel, Mark and Freitag, Markus and Gowda, Thamme and
                   Grundkiewicz, Roman and Haddow, Barry and
                   Karpinska, Marzena and Koehn, Philipp and
                   Lakougna, Howard and Lundin, Jessica and
                   Monz, Christof and Murray, Kenton and Nagata, Masaaki and
                   Perrella, Stefano and Proietti, Lorenzo and
                   Popel, Martin and Popovi{\'c}, Maja and Riley, Parker and
                   Shmatova, Mariya and Steingr{\'i}msson, Steinth{\'o}r and
                   Yankovskaya, Lisa and Zouhar, Vil{\'e}m},
  booktitle =     {Proceedings of the Tenth Conference on Machine
                   Translation},
  editor =        {Haddow, Barry and Kocmi, Tom and Koehn, Philipp and
                   Monz, Christof},
  month =         nov,
  pages =         {355--413},
  publisher =     {Association for Computational Linguistics},
  title =         {Findings of the {WMT}25 General Machine Translation
                   Shared Task: Time to Stop Evaluating on Easy Test
                   Sets},
  year =          {2025},
  doi =           {10.18653/v1/2025.wmt-1.22},
  isbn =          {979-8-89176-341-8},
  url =           {https://aclanthology.org/2025.wmt-1.22/},
}

@inproceedings{barrault-etal-2019-findings,
  address =       {Florence, Italy},
  author =        {Barrault, Lo{\"i}c and Bojar, Ond{\v{r}}ej and
                   Costa-juss{\`a}, Marta R. and Federmann, Christian and
                   Fishel, Mark and Graham, Yvette and Haddow, Barry and
                   Huck, Matthias and Koehn, Philipp and
                   Malmasi, Shervin and Monz, Christof and
                   M{\"u}ller, Mathias and Pal, Santanu and Post, Matt and
                   Zampieri, Marcos},
  booktitle =     {Proceedings of the Fourth Conference on Machine
                   Translation (Volume 2: Shared Task Papers, Day 1)},
  editor =        {Bojar, Ond{\v{r}}ej and Chatterjee, Rajen and
                   Federmann, Christian and Fishel, Mark and
                   Graham, Yvette and Haddow, Barry and Huck, Matthias and
                   Yepes, Antonio Jimeno and Koehn, Philipp and
                   Martins, Andr{\'e} and Monz, Christof and
                   Negri, Matteo and N{\'e}v{\'e}ol, Aur{\'e}lie and
                   Neves, Mariana and Post, Matt and Turchi, Marco and
                   Verspoor, Karin},
  month =         aug,
  pages =         {1--61},
  publisher =     {Association for Computational Linguistics},
  title =         {Findings of the 2019 Conference on Machine
                   Translation ({WMT}19)},
  year =          {2019},
  doi =           {10.18653/v1/W19-5301},
  url =           {https://aclanthology.org/W19-5301/},
}

@inproceedings{kocmi-etal-2023-findings,
  address =       {Singapore},
  author =        {Kocmi, Tom and Avramidis, Eleftherios and
                   Bawden, Rachel and Bojar, Ond{\v{r}}ej and
                   Dvorkovich, Anton and Federmann, Christian and
                   Fishel, Mark and Freitag, Markus and Gowda, Thamme and
                   Grundkiewicz, Roman and Haddow, Barry and
                   Koehn, Philipp and Marie, Benjamin and Monz, Christof and
                   Morishita, Makoto and Murray, Kenton and
                   Nagata, Masaaki and Nakazawa, Toshiaki and
                   Popel, Martin and Popovi{\'c}, Maja and
                   Shmatova, Mariya and Suzuki, Jun},
  booktitle =     {Proceedings of the Eighth Conference on Machine
                   Translation},
  editor =        {Koehn, Philipp and Haddow, Barry and Kocmi, Tom and
                   Monz, Christof},
  month =         dec,
  pages =         {1--42},
  publisher =     {Association for Computational Linguistics},
  title =         {Findings of the 2023 Conference on Machine
                   Translation ({WMT}23): {LLM}s Are Here but Not Quite
                   There Yet},
  year =          {2023},
  doi =           {10.18653/v1/2023.wmt-1.1},
  url =           {https://aclanthology.org/2023.wmt-1.1/},
}

@inproceedings{rei-etal-2022-comet,
  address =       {Abu Dhabi, United Arab Emirates (Hybrid)},
  author =        {Rei, Ricardo and C. de Souza, Jos{\'e} G. and
                   Alves, Duarte and Zerva, Chrysoula and Farinha, Ana C and
                   Glushkova, Taisiya and Lavie, Alon and Coheur, Luisa and
                   Martins, Andr{\'e} F. T.},
  booktitle =     {Proceedings of the Seventh Conference on Machine
                   Translation (WMT)},
  editor =        {Koehn, Philipp and Barrault, Lo{\"i}c and
                   Bojar, Ond{\v{r}}ej and Bougares, Fethi and
                   Chatterjee, Rajen and Costa-juss{\`a}, Marta R. and
                   Federmann, Christian and Fishel, Mark and
                   Fraser, Alexander and Freitag, Markus and
                   Graham, Yvette and Grundkiewicz, Roman and
                   Guzman, Paco and Haddow, Barry and Huck, Matthias and
                   Jimeno Yepes, Antonio and Kocmi, Tom and
                   Martins, Andr{\'e} and Morishita, Makoto and
                   Monz, Christof and Nagata, Masaaki and
                   Nakazawa, Toshiaki and Negri, Matteo and
                   N{\'e}v{\'e}ol, Aur{\'e}lie and Neves, Mariana and
                   Popel, Martin and Turchi, Marco and Zampieri, Marcos},
  month =         dec,
  pages =         {578--585},
  publisher =     {Association for Computational Linguistics},
  title =         {{COMET}-22: Unbabel-{IST} 2022 Submission for the
                   Metrics Shared Task},
  year =          {2022},
  doi =           {10.18653/v1/2022.wmt-1.52},
  url =           {https://aclanthology.org/2022.wmt-1.52/},
}

@inproceedings{juraska-etal-2024-metricx,
  address =       {Miami, Florida, USA},
  author =        {Juraska, Juraj and Deutsch, Daniel and
                   Finkelstein, Mara and Freitag, Markus},
  booktitle =     {Proceedings of the Ninth Conference on Machine
                   Translation},
  editor =        {Haddow, Barry and Kocmi, Tom and Koehn, Philipp and
                   Monz, Christof},
  month =         nov,
  pages =         {492--504},
  publisher =     {Association for Computational Linguistics},
  title =         {{M}etric{X}-24: The {G}oogle Submission to the {WMT}
                   2024 Metrics Shared Task},
  year =          {2024},
  doi =           {10.18653/v1/2024.wmt-1.35},
  url =           {https://aclanthology.org/2024.wmt-1.35/},
}

@inproceedings{hu-etal-2022-lora,
  author =        {Hu, Edward J. and Shen, Yelong and Wallis, Phillip and
                   Allen-Zhu, Zeyuan and Li, Yuanzhi and Wang, Shean and
                   Wang, Lu and Chen, Weizhu},
  booktitle =     {International Conference on Learning Representations
                   (ICLR)},
  title =         {{L}o{RA}: Low-Rank Adaptation of Large Language
                   Models},
  year =          {2022},
  url =           {https://arxiv.org/abs/2106.09685},
}

@inproceedings{kingma-ba-2015-adam,
  author =        {Kingma, Diederik P. and Ba, Jimmy},
  booktitle =     {International Conference on Learning Representations},
  title =         {Adam: A Method for Stochastic Optimization},
  year =          {2015},
  url =           {https://arxiv.org/abs/1412.6980},
}

@inproceedings{furlanello-etal-2018-born,
  author =        {Furlanello, Tommaso and Lipton, Zachary and
                   Tschannen, Michael and Itti, Laurent and
                   Anandkumar, Anima},
  booktitle =     {Proceedings of the 35th International Conference on
                   Machine Learning (ICML)},
  pages =         {1607--1616},
  title =         {Born-Again Neural Networks},
  volume =        {80},
  year =          {2018},
  url =           {https://proceedings.mlr.press/v80/furlanello18a.html},
}

@inproceedings{zheng-etal-2019-simpler,
  address =       {Hong Kong, China},
  author =        {Zheng, Baigong and Zheng, Renjie and Ma, Mingbo and
                   Huang, Liang},
  booktitle =     {Proceedings of the 2019 Conference on Empirical
                   Methods in Natural Language Processing and the 9th
                   International Joint Conference on Natural Language
                   Processing (EMNLP-IJCNLP)},
  editor =        {Inui, Kentaro and Jiang, Jing and Ng, Vincent and
                   Wan, Xiaojun},
  month =         nov,
  pages =         {1349--1354},
  publisher =     {Association for Computational Linguistics},
  title =         {Simpler and Faster Learning of Adaptive Policies for
                   Simultaneous Translation},
  year =          {2019},
  doi =           {10.18653/v1/D19-1137},
  url =           {https://aclanthology.org/D19-1137/},
}

@inproceedings{ma-etal-2020-monotonic,
  author =        {Ma, Xutai and Pino, Juan Miguel and Cross, James and
                   Puzon, Liezl and Gu, Jiatao},
  booktitle =     {International Conference on Learning Representations
                   (ICLR)},
  title =         {Monotonic Multihead Attention},
  year =          {2020},
  url =           {https://openreview.net/forum?id=Hyg96gBKPS},
}

\appendix
%%%%%%%%%%%%%%%%%%%%%%%%%%%%%%%%%%%%%%%%%%%%%%%%%%
\section*{Appendices}

\section{Stable-prefix construction}
\label{sec:prefix-algorithms}

\textbf{Input:} source sentence $s=(w_1,\ldots,w_n)$; translation model $t$\\
\textbf{Output:} prefix training pairs $D$

\begin{enumerate}
\item Set $T \leftarrow t(s)$, $m \leftarrow 0$, and $D \leftarrow \varnothing$.
\item For $i=1,\ldots,n$:
  \begin{enumerate}
\item Set $s_i \leftarrow (w_1,\ldots,w_i)$ and $P_i \leftarrow t(s_i)$.
\item Set $\ell_i \leftarrow |\operatorname{lcp}(P_i,T)|$.
\item Enforce monotonicity: $m \leftarrow \max(m,\ell_i)$.
\item Add $(s_i,T_{1..m})$ to $D$.
  \end{enumerate}
\item Return $D$.
\end{enumerate}

Here $\operatorname{lcp}$ is the longest common target prefix. The running maximum ensures that the committed target length never decreases even as more source words arrive.

\section{Reproducibility details}
\label{sec:reproducibility}

\paragraph{Average Lagging.} We use the standard Average Lagging of \citet{ma-etal-2019-stacl}. Latency is measured in source-word units. For a sentence with $n$ source words and $T$ committed target units, let $g_i$ be the number of source words read when target unit $i$ is committed, $\gamma=T/n$, and $\tau$ the first target position committed after all $n$ source words have been read (or $T$ if this never occurs). We compute
\begin{equation}
\operatorname{AL}=\frac{1}{\tau}\sum_{i=1}^{\tau}\left(g_i-\frac{i-1}{\gamma}\right).
\end{equation}
The target units are whitespace-delimited words for German, and characters for Japanese and Chinese.

\paragraph{Attainable-envelope AUC.} Each dataset--language panel uses one shared latency interval $[L,U]$, where $L$ and $U$ are respectively the minimum and maximum AL over the union of all operating points from all compared sweeps in that panel. We evaluate 101 equally spaced budgets $b_j=L+j(U-L)/100$. For COMET, let $S_s$ be sweep $s$'s set of (AL, quality) points, let $q_{\mathrm{worst}}$ be the lowest COMET value anywhere in the panel, and define
\begin{equation}
\begin{aligned}
\mathcal{Q}_s(b)&=\{q:(a,q)\in S_s,\ a\le b\},\\
F_s(b)&=\begin{cases}
\max \mathcal{Q}_s(b),&\mathcal{Q}_s(b)\ne\varnothing,\\
q_{\mathrm{worst}},&\text{otherwise},
\end{cases}\\
\operatorname{AUC}_s&=\frac{1}{101}\sum_{j=0}^{100}F_s(b_j).
\end{aligned}
\end{equation}
Thus every sweep has the same domain and denominator, including below its minimum attainable latency. For MetricX, where lower is better, we replace both maxima by minima and use the panel's highest error as $q_{\mathrm{worst}}$. The five sweeps are \code{base} with wait-$k$, suffix deletion, or confidence, and \code{prefix} and \code{continuation} with confidence.

\paragraph{Optimization.} The evaluated adapters use LoRA rank 16, scaling factor 32, dropout 0.05, and no bias, applied to the attention $q$, $k$, $v$, and output projections and the MLP gate, up, and down projections. The continuation stage initializes the selected prefix adapter and keeps the same adapter trainable. Both stages use fused AdamW, a constant learning rate of $5\times10^{-6}$ without warmup, gradient clipping at 1.0, and an effective batch size of 128.

\paragraph{Corpus parsing and sampling.} The synthetic corpus was constructed from a nominal 20 million source sentences per language direction. However, very little of the corpus is actually used in training because training stops early when the validation loss stops improving. Table~\ref{tab:training-counts} shows the number of updates and total records processed for each language direction.

\begin{table}[t]
\centering
\small
\begin{tabular}{@{}lcc@{}}
\toprule
Direction & \code{prefix} & \code{continuation} \\
\midrule
EN$\rightarrow$DE & 2,000 / 4,500 (576k) & 3,500 / 6,000 (768k) \\
EN$\rightarrow$JA & 1,000 / 3,500 (448k) & 10,500 / 13,000 (1.664M) \\
EN$\rightarrow$ZH & 2,000 / 4,500 (576k) & 8,500 / 11,000 (1.408M) \\
\bottomrule
\end{tabular}
\caption{Selected update / final update evaluated before stopping; parentheses give the total number of optimized records through the final update (updates multiplied by effective batch size 128).}
\label{tab:training-counts}
\end{table}

\section{LLM Prompt Templates}
\label{sec:llm-prompt-templates}

All prompts use a single force-decode turn rendered through the Qwen3 chat template (thinking disabled): a system message, one user message carrying the instruction and all source observed so far, and an open assistant turn seeded with the committed target (empty on the first step) that the model continues. The instruction takes one of two forms depending on whether the sentence is yet complete. The examples below use the running source and committed translation example from the LLM prompting in Section \ref{sec:LLM Prompting}:

\subsection{Prompts}

\paragraph{Initial translation} With no target committed yet, the single user turn translates the source observed so far and the model writes from an empty prefix: 

{\scriptsize
\begin{verbatim}
[system]    You are a professional simultaneous
            translator from English to German.
[user]      An initial English segment has
            arrived, but the sentence is not
            complete. Translate only what is
            supported so far. Output ONLY the
            German translation.
            English: Scientists discovered a new
-->         Wissenschaftler entdeckten eine neue
\end{verbatim}
}

\paragraph{Continuation translation} Once a target prefix has been committed, the model is restarted with all source observed so far and the committed target supplied as a forced prefix; free decoding then emits only the new continuation:

{\scriptsize
\begin{verbatim}
[system]    You are a professional simultaneous
            translator from English to German.
[user]      An initial English segment has
            arrived, but the sentence is not
            complete. Translate only what is
            supported so far. Output ONLY the
            German translation.
            English: Scientists discovered a
            new species
[forced]    Wissenschaftler entdeckten eine neue
-->         Art
\end{verbatim}
}

\paragraph{Final translation} When the full sentence has arrived, the instruction switches to its final form: the user turn carries the complete source and the committed target is again forced, so the model produces only the remaining continuation: 

{\scriptsize
\begin{verbatim}
[system]    You are a professional simultaneous
            translator from English to German.
[user]      The final English segment has
            arrived; the sentence is now
            complete. Keep the existing German
            unchanged and output ONLY the new
            continuation needed to complete the
            translation.
            English: Scientists discovered a new
            species in the rainforest.
[forced]    Wissenschaftler entdeckten eine neue
            Art
-->         im Regenwald.
\end{verbatim}
}

\section{MetricX evaluation}
\label{sec:metricx}

\begin{figure*}[t]
\centering
\textbf{FLEURS}\\[1pt]
\begin{minipage}{0.32\linewidth}\centering
\textbf{\footnotesize EN$\rightarrow$DE}\\[2pt]
  \includegraphics[width=\linewidth]{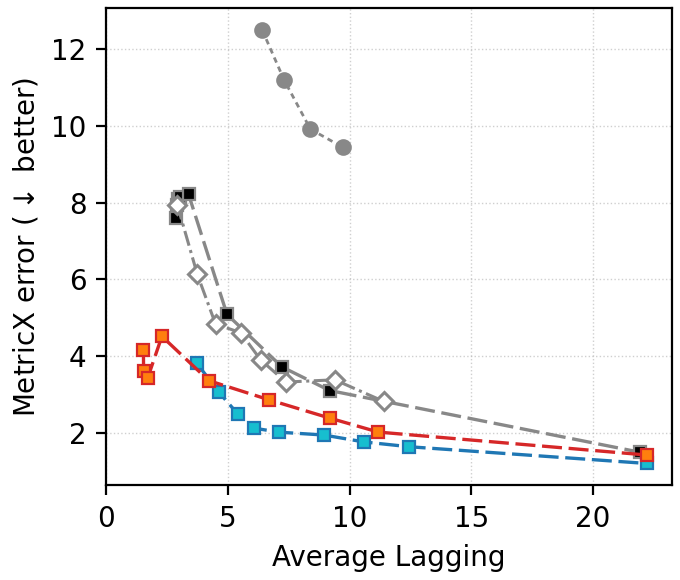}
\end{minipage}\hfill
\begin{minipage}{0.32\linewidth}\centering
\textbf{\footnotesize EN$\rightarrow$JA}\\[2pt]
  \includegraphics[width=\linewidth]{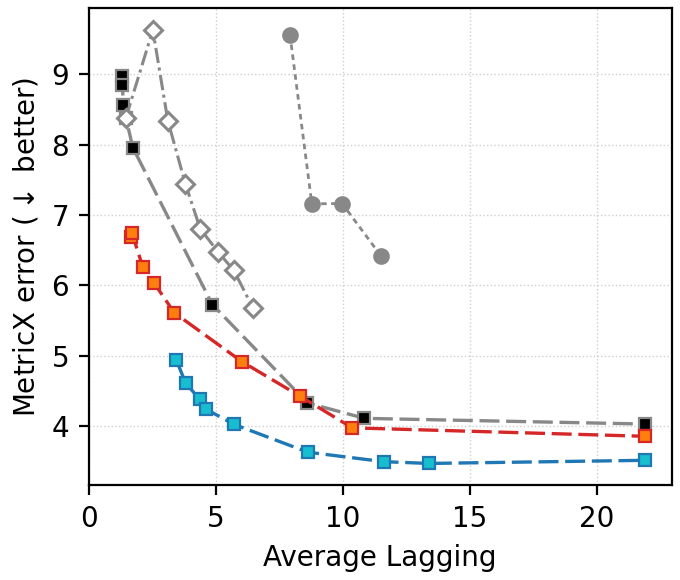}
\end{minipage}\hfill
\begin{minipage}{0.32\linewidth}\centering
\textbf{\footnotesize EN$\rightarrow$ZH}\\[2pt]
  \includegraphics[width=\linewidth]{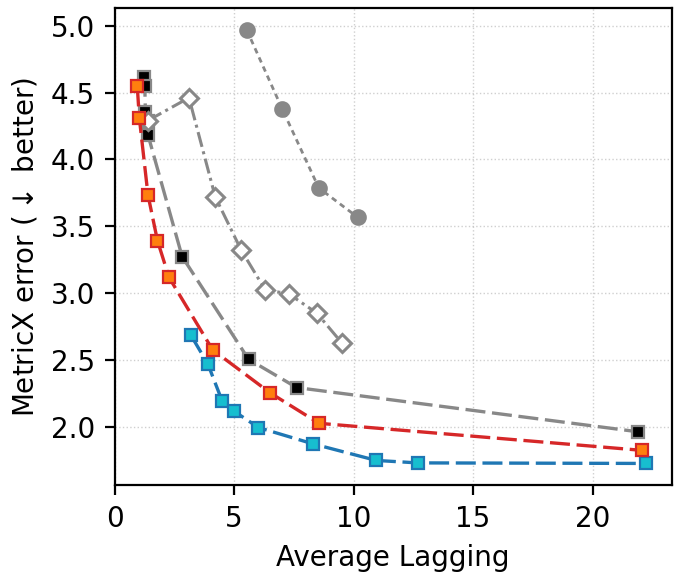}
\end{minipage}

\vspace{3pt}
\textbf{WMT24{+}{+}}\\[1pt]
\begin{minipage}{0.32\linewidth}\centering
  \includegraphics[width=\linewidth]{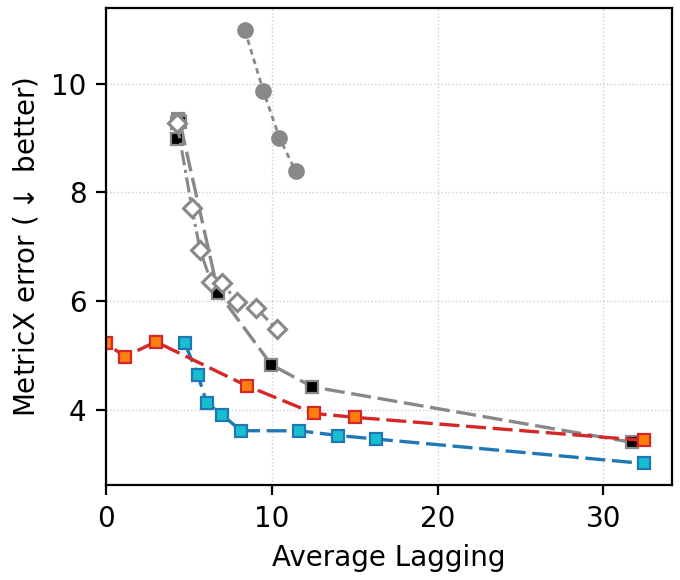}
\end{minipage}\hfill
\begin{minipage}{0.32\linewidth}\centering
  \includegraphics[width=\linewidth]{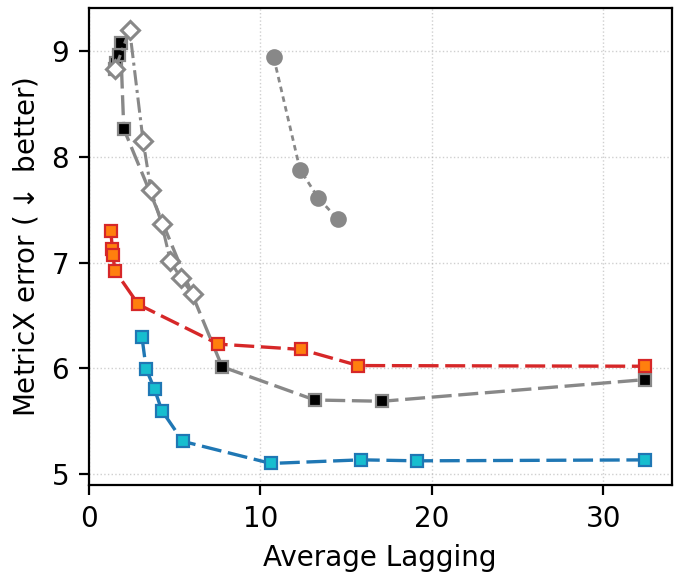}
\end{minipage}\hfill
\begin{minipage}{0.32\linewidth}\centering
  \includegraphics[width=\linewidth]{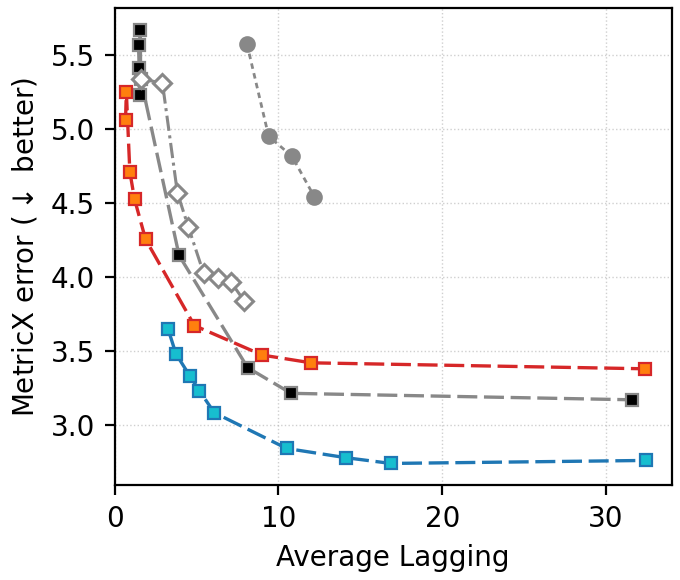}
\end{minipage}

\vspace{3pt}
\textbf{CoVoST2}\\[1pt]
\begin{minipage}{0.32\linewidth}\centering
  \includegraphics[width=\linewidth]{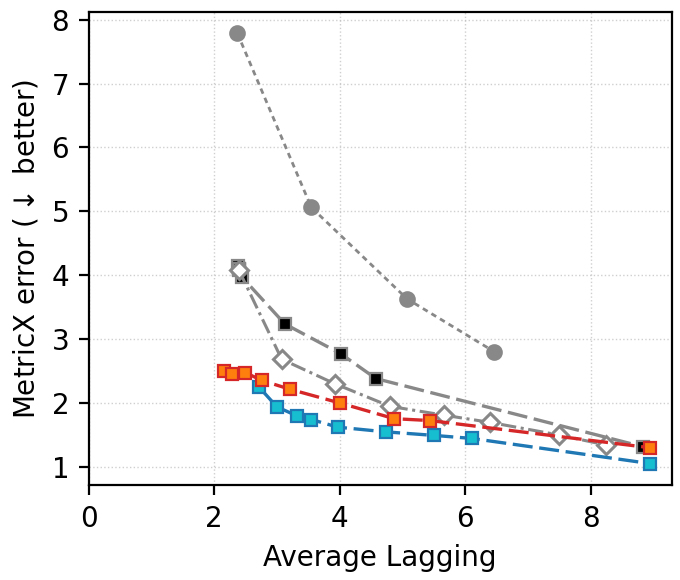}
\end{minipage}\hfill
\begin{minipage}{0.32\linewidth}\centering
  \includegraphics[width=\linewidth]{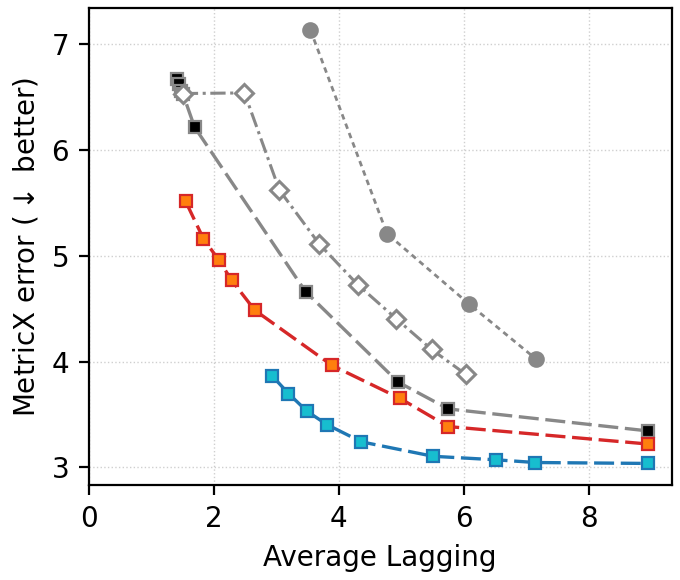}
\end{minipage}\hfill
\begin{minipage}{0.32\linewidth}\centering
  \includegraphics[width=\linewidth]{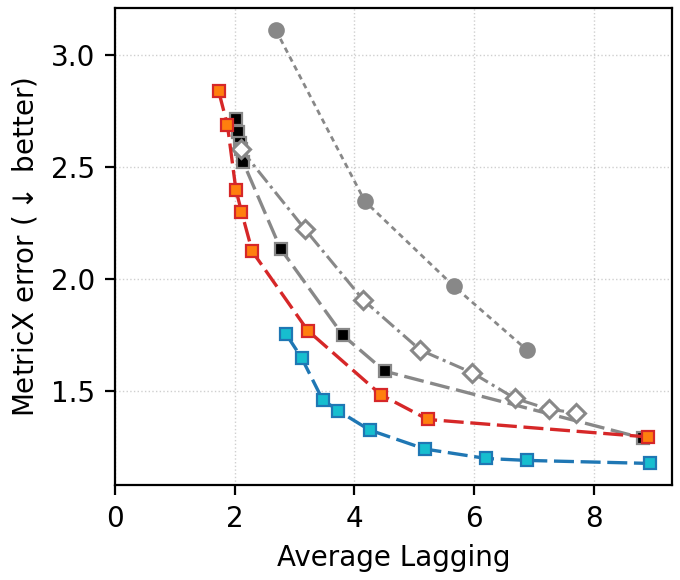}
\end{minipage}

\vspace{2pt}
\includegraphics[width=0.7\textwidth]{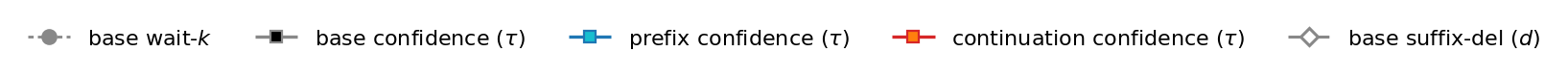}
\caption{MetricX version of Figure~\ref{fig:sched-robust}, with the same five sweeps (MetricX error, \emph{lower is better}).}
\label{fig:sched-robust-metricx}
\end{figure*}

As a metric-robustness check we re-score the FLEURS, WMT24++, and CoVoST~2 evaluations with MetricX, a
reference-based error metric that uses a range of 0-25 (lower is better). Figure~\ref{fig:sched-robust-metricx} is the
MetricX copy of the quality--latency mechanism comparison in Figure~\ref{fig:sched-robust}. Table~\ref{tab:sched-robust-auc-metricx} reports the corresponding MetricX frontier AUC for each sweep.
The qualitative ordering carries over between metrics: \code{prefix} is strongest at higher
latency, while \code{continuation} reaches the lowest-latency band.

\begin{table}[t]
\centering
\scriptsize
\caption{MetricX-error quality--latency frontier AUC corresponding to Table~\ref{tab:sched-robust-auc} (lower is better), for EN$\rightarrow$\{DE,JA,ZH\}. Each column scores one sweep on the shared latency interval for that dataset--language panel; best per row in \textbf{bold}.}
\label{tab:sched-robust-auc-metricx}
\setlength{\tabcolsep}{3pt}
\begin{tabular}{@{}lrrrrr@{}}
\toprule
 & \multicolumn{3}{c}{\code{base}} & \multicolumn{2}{c}{ft.\,+\,conf.} \\
\cmidrule(lr){2-4}\cmidrule(l){5-6}
 & wait-$k$ & suf.\,del. & conf. & \code{prefix} & \code{cont.} \\
\midrule
% BEGIN SCHED_ROBUST_METRICX_AUC
\multicolumn{6}{@{}l}{\textbf{FLEURS}} \\
DE & 10.4 & 4.2 & 4.5 & 3.1 & \textbf{2.5} \\
JA & 7.7 & 6.2 & 5.1 & \textbf{4.3} & 4.6 \\
ZH & 4.0 & 3.0 & 2.6 & \textbf{2.2} & 2.3 \\
\midrule
\multicolumn{6}{@{}l}{\textbf{WMT24++}} \\
DE & 9.3 & 6.7 & 6.1 & 5.0 & \textbf{4.3} \\
JA & 8.1 & 6.9 & 6.3 & \textbf{5.4} & 6.2 \\
ZH & 4.9 & 4.0 & 3.6 & \textbf{3.1} & 3.6 \\
\midrule
\multicolumn{6}{@{}l}{\textbf{CoVoST2}} \\
DE & 4.5 & 2.4 & 2.9 & 2.1 & \textbf{1.9} \\
JA & 5.7 & 4.8 & 4.5 & \textbf{4.0} & \textbf{4.0} \\
ZH & 2.3 & 1.9 & 1.8 & \textbf{1.6} & \textbf{1.6} \\
% END SCHED_ROBUST_METRICX_AUC
\bottomrule
\end{tabular}
\end{table}

\end{document}